\documentclass{article} 
\usepackage{iclr2027_conference,times}
\iclrfinalcopy

\title{Fast Generalized Neural Tangent Kernel Statistics via Trace Estimation}

\author{
James Hazelden$^{1,2}$, Balaaji Reddy Nagireddy$^{1}$, \& Eric Shea-Brown$^{1,2}$\\
$^1$ Applied Mathematics, University of Washington\\
$^2$ The Allen Institute\\
\text{  } Seattle, WA, USA\\
\texttt{\{jhazelde,balaaji,etsb\}@uw.edu}
}

\date{September 2026}
\usepackage{amsmath, amsfonts, amssymb}
\usepackage{float}
\usepackage{algorithm}
\usepackage{algpseudocode}
\usepackage{booktabs,tabularx,array}
\newcolumntype{Y}{>{\centering\arraybackslash}X}
\usepackage{graphicx} 
\usepackage{url}
\usepackage{verbatim}

\renewcommand{\d}{\text{d}}

\newcommand{\R}{\mathbb{R}}

\newcommand{\tr}{\text{tr}}
\newcommand{\vjp}{\texttt{vjp}}
\newcommand{\jvp}{\texttt{jvp}}

\newcommand{\NTK}{\texttt{NTK}}

\newcommand{\EcommentDONE}[1]{} 
\newcommand{\BcommentDONE}[1]{}

\algrenewcommand\algorithmicrequire{\textbf{Input:}}
\algrenewcommand\algorithmicensure{\textbf{Output:}}
\newcommand{\sectn}[1]{[§\ref{sec:#1}]}
\newcommand{\todo}[1]{}

\makeatletter
\newcommand{\propositionname}{Proposition}
\newenvironment{proposition}[1][]%
{%
  \par\medskip
  \noindent
  \textbf{\propositionname~1\if\relax\detokenize{#1}\relax\else\ (\textit{#1})\fi.}%
  \quad\ignorespaces
}%
{%
  \par\medskip
}
\makeatother

\begin{document}
\maketitle

\begin{abstract}
The empirical state-space Neural Tangent Kernel (NTK) describes the local learning geometry of a finite-width neural network, but computing it explicitly is almost always impractical in terms of computation and memory costs. Here, we show that many useful NTK statistics that characterize, for example, the dimensionality of learned updates or how two models or learning rules relate, can instead be efficiently approximated to very high accuracy via matrix-free products using randomized trace estimation. Namely, we use Hutch++ to estimate the NTK trace, Frobenius norm, effective rank, and alignment. Furthermore, we show that the positive-semidefinite structure of the NTK yields \emph{one-sided} estimators that require only forward- or reverse-mode automatic differentiation. We validate these estimators across MLPs, recurrent GRUs, and a natural-language Transformer with up to 410 million parameters, in which the state-space contains high-dimensional four-tensors. We demonstrate orders-of-magnitude speedups, with the fastest estimator in a given application depending on the ratio of parameter and state dimensions. Equipped with these estimators, we examine rich and lazy RNN training using hidden-state NTK alignment and use NTK alignment as a regularizer for data-scarce knowledge distillation. We find that this regularization can modestly improve generalization, especially in very data-scarce settings. Together, these results suggest state-space NTK diagnostics are practical even at large scales.
\end{abstract}

\vspace{-.6cm}

\todo{note somewhere hidden state NTK and output have zero diff in runtime here}


\section{Introduction}

\begin{figure}[t]
    \centering
    \includegraphics[width=.94\linewidth]{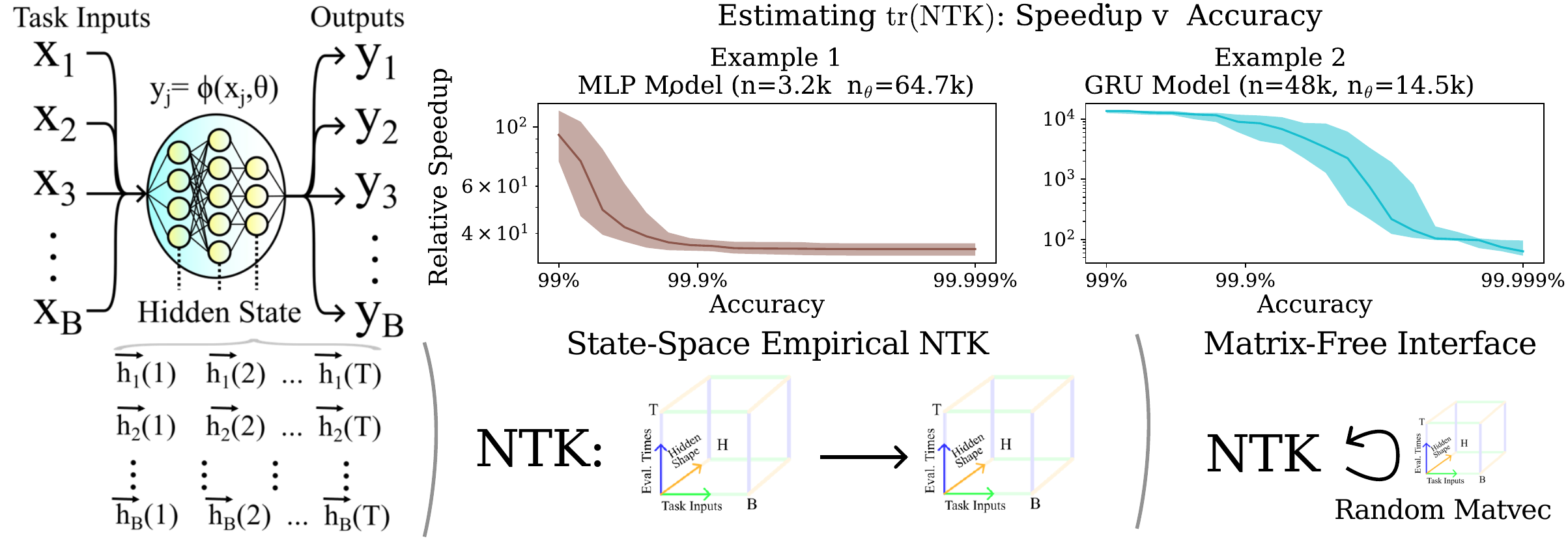}
    
    \caption{\textbf{Paper Overview:  The empirical state-space NTK and trace estimation.} The NTK acts on the full model state, which can form a high-dimensional tensor, and is accessed here only through randomized matrix-vector products (matvecs). For representative MLP and GRU examples (top right, see \sectn{numres}), the fastest of the three estimators discussed achieves 99\% accuracy about \(80\) and \(10{,}000\) times faster than exact trace evaluation, respectively. Curves show medians over 50 runs with interquartile ranges. \vspace{-.5em}}
    \label{fig:summary}
\end{figure}

The Neural Tangent Kernel (NTK) is a linear operator describing the local change in a model's state under gradient descent (GD) \citep{jacot2018neural}. Classically, the NTK is defined on model outputs and is primarily studied via infinite-width limits \citep{mei2018mean, pmlr-v139-yang21c, bordelon2020spectrum}. Here, we use the more general notion of an NTK defined on a prescribed state space and focus on computational tools that enable its quantification. For example, for an MLP, the state might contain activations at every layer over an empirical batch of inputs (Figure~\ref{fig:summary}). The resulting state-space NTK can describe how \textit{all} selected state variables evolve during training, not just the readout~\citep{hazelden2026globalempiricalntkselfreferential, liu2024how}. State-space NTK statistics of interest include the trace, Frobenius norm, alignment, and effective rank, which respectively quantify overall tangent sensitivity, kernel magnitude and drift, and concentration of learning onto low-dimensional update directions~\citep{xiang2025neuraltangentknowledgedistillation, sharma2025relationship, liu2024how, shan2022theoryneuraltangentkernel, DBLP:journals/corr/abs-1906-08034, atanasov2021neuralnetworkskernellearners, kumar2024grokkingtransitionlazyrich, internalDynamicsNTK2025, hazelden2026globalempiricalntkselfreferential, murray2023characterizing}. These quantities can all be written using traces of the NTK or products of NTKs~\sectn{terms}.

The obstacle is computational. Explicitly constructing a finite-width NTK scales poorly with the dimension of the state, becoming especially prohibitive when it includes batches, recurrent timesteps, hidden units, Transformer blocks, or tokens. An RNN with 100 hidden units, evaluated on 100 inputs for 100 timesteps, already has a state with one million entries and an NTK with \(10^{12}\) entries. We instead treat the NTK as a \textit{matrix-free} operator and use randomized trace estimation to recover statistics. In particular, we apply Hutch++ \citep{meyer2021hutchoptimalstochastictrace} to the NTK and its products, as well as deriving one-sided estimators relying only on forward- or reverse-mode automatic differentiation. 

Across an MLP, a recurrent GRU, and the pretrained Pythia natural language Transformer~\citep{biderman2023pythiasuiteanalyzinglarge}, these estimators recover near-exact NTK traces at a small fraction of the cost of exact state-basis evaluation. In the case of Pythia-410M, for example, the fastest estimator reaches 99\% accuracy about \(13{,}500\) times faster than exact evaluation \sectn{transformer}. The same machinery efficiently estimates NTK norm, alignment, and effective rank, making these quantities practical as training-time diagnostics. We use this capability to track rich versus lazy recurrent learning at high temporal resolution and to regularize data-limited student networks toward a teacher's tangent geometry, improving held-out generalization. 

Figure~\ref{fig:summary} provides a summary. Our \textit{primary contributions} are
\begin{enumerate}
    \item \textbf{Matrix-free NTK statistics.} We formulate fast estimation of finite-width NTK trace, Frobenius norm, alignment, and effective rank using Hutch++ and matrix-free NTK products and quantify their runtime-accuracy performance on MLP, GRU, and Pythia Transformer models up to 410M parameters \sectn{methods}.
    \item \textbf{One-sided trace estimation and scaling.} We derive reverse-only and forward-only NTK trace estimators and validate them on the same class of models \sectn{numres}.
    \item \textbf{Practical applications.} We use the estimators to study rich and lazy recurrent learning across stages of learning, and to impose teacher-student NTK alignment as a data-scarce regularizer, producing reproducible held-out generalization gains \sectn{applications}.
\end{enumerate}

\section{Related Work}

\paragraph{The Neural Tangent Kernel}
The Neural Tangent Kernel (NTK) was introduced by Jacot et al.~\cite{jacot2018neural} as the kernel governing the first-order evolution of network outputs under gradient descent. Subsequent work has developed two broad, complementary directions. Infinite-width theory studies regimes in which the NTK converges to a deterministic kernel and training approaches kernel regression, with extensions to general architectures, including recurrent and equilibrium models~\citep{lee2019wide,yang2020tensorprogramsII,alemohammad2020recurrent,feng2023deqntk}. In parallel, computational work has developed efficient implementations and approximations of infinite- and finite-width NTKs, using analytic recursions, structured Jacobian contractions, output-space approximations, and randomized sketches~\citep{novak2019neuraltangentsfasteasy,arora2019exact,novak2022fastfinitewidthneural,mohamadi2023fast,zandieh2021scalingneuraltangentkernels}. The canonical NTK acts on model outputs. Here, instead, we consider the empirical \emph{state-space NTK}: the state $h$ collects the model's internal variables across examples, layers, or time, and $\mathrm{NTK}_S = J_hJ_h^\top$ acts directly on this tensor space. Once a loss specifies the state-space error signal, this operator closes the first-order learning dynamics at the level of the model state, exposing representation dynamics that are absent from an output-only view~\citep{hazelden2026globalempiricalntkselfreferential}. 

\paragraph{Operators in Deep Learning}
More broadly, our approach follows a growing trend within deep learning that seeks to represent underlying very large, structured operators and tensors through their \emph{action}, rather than explicitly materializing these objects. Curvlinops applies this principle to Hessian, Fisher/GGN, and structured curvature matrices, exposing them through a common linear-operator interface~\citep{dangel2025positioncurvaturematricesdemocratized}; this perspective is closely connected to randomized numerical linear algebra, where low-rank structure and matrix statistics are recovered from matrix-vector products~\cite{martinsson2020randomized}. Related work by Townsend and collaborators studies recovery of infinite-dimensional operators and Green's functions from operator queries, including settings in which the adjoint is unavailable~\citep{boulle2024operator}. Trace estimation provides a particularly mature example: Hutchinson's estimator~\citep{hutchinson1989stochastic} has been refined by Hutch++~\citep{meyer2021hutchoptimalstochastictrace}, adaptive and Nystr\"om variants~\citep{persson2022improved}, XTrace~\citep{xtrace}, and extensions to continuous implicit operators~\citep{zvonek2025conthutch}. 

Finally, prior work, in the context of dynamical operator decompositions of the empirical NTK, briefly discussed Hutch++ for estimating the NTK trace quantities ~\citep{hazelden2026globalempiricalntkselfreferential}. The present paper explains the underlying computational and mathematical structure, derives and analyzes new one-sided estimators, quantifies runtime improvements across varied models, including pretrained natural language transformers, and studies new applications, including the gradual emergence of rich vs lazy representations and NTK-informed knowledge distillation.

\section{NTK Background}

\label{sec:background}

\paragraph{NTK Definition} The empirical state-space Neural Tangent Kernel (NTK) is a linear operator defining the first-order dependence of parameterized state variables on their parameters. Crucially, it defines the geometry of gradient descent (GD) learning at each iteration to first order~\citep{jacot2018neural,hazelden2026globalempiricalntkselfreferential}. Consider a (typically tensor-valued) state dependent on parameters \(\theta \in P\), \(h = h(\theta) \in H\). 

Throughout, we assume \(\text{vec}(h) \in \R^n\), the state, reduced to a vector, has \(n\) entries. Next, define the parameter Jacobian of this mapping \(J_\theta := \frac{\d h}{\d \theta}\), which may be tensor-valued in general. Then, we can define two operations \(\vjp : H \rightarrow P, \jvp : P \rightarrow H\) (vector-Jacobian and Jacobian-vector product, respectively). Specifically, given any tangent tensor \(v \in H\) or \(p \in P\) we define \(\vjp(v) := J_\theta^T v; \,  \jvp(p) := J_\theta p\). From this, \(\NTK : H \rightarrow H\) is a linear operator defined by  
\begin{align}
    \NTK (v) := \jvp(\vjp(v)) = J_\theta J_\theta^T v \quad \text{ for any } v \in H
\end{align}
Importantly, \(J_\theta\) is never explicitly constructed since this matrix can be prohibitively large. Performing \(\vjp\) uses \textit{reverse-mode} automatic differentiation (AD), and \(\jvp\) uses \textit{forward-mode} AD~\citep{paszke2019pytorchimperativestylehighperformance}. We note that sometimes the left and right Jacobians above differ (the cross-NTK), \(J_\theta(x) J_\theta(x')^\top\), for different inputs \(x, x'\)~\citep{lee2019wide}. For simplicity, we assume the structure above, but Hutch++ (Algorithm~\ref{alg:hutchpp_ntk}) applies to this non-symmetric case immediately, and the one-sided estimator (Algorithm~\ref{alg:oneside_trace}) can be tweaked to apply, as detailed in Proposition 1 in the Appendix~\ref{sec:aponeside}.

%

\subsection{Statistics Derived From the Trace}

\label{sec:terms}

 Here, we discuss a few statistics derived from the trace and motivated applications: the trace, alignment, effective rank, and Frobenius norm. Each can be estimated with Hutch++, as well as with one-sided estimators, as discussed in Appendix~\ref{sec:aponeside}.

 \paragraph{Trace}

 The trace itself is of interest, since \(\tr(\NTK) = \|J_\theta\|_F^2\), describing the overall sensitivity of the state to the parameters. Such a quantity, capturing total sensitivity, could be used, for example, to detect sudden changes in learning, such as bifurcations, known as \textit{critical slowing down}, or to detect emergent stiff parameters~\citep{kramer1985stabilization,anderson2023sloppy,chen2026variancesensitivitybaseddimensionsbrain}.
 
\paragraph{Alignment Between NTKs}

After a model is trained on a task, how much the NTK changes can quantify rich and lazy GD training regimes \citep{liu2024how, shan2022theoryneuraltangentkernel, DBLP:journals/corr/abs-1906-08034, atanasov2021neuralnetworkskernellearners, kumar2024grokkingtransitionlazyrich}. This can be measured by how much spans overlap, specifically, given two NTKs with the same underlying state-space, \(\NTK_1, \NTK_2 : H \rightarrow H\), their alignment is  
\begin{align}
    \label{eqn:align}
    \cos(\NTK_1, \NTK_2) := \frac{\tr(\NTK_1^T \NTK_2)}{\|\NTK_1\|_F \|\NTK_2\|_F} \in [0,1]
\end{align}

\paragraph{Effective Rank} 

The NTK effective rank can quantify how constrained and potentially low-dimensional GD updates are \citep{internalDynamicsNTK2025, hazelden2026globalempiricalntkselfreferential, murray2023characterizing}. This can be expressed through the \textit{participation ratio} (PR),
\begin{align}
    r_{\mathrm{eff}}(\NTK) := \frac{\tr(\NTK)^2}{\tr(\NTK \cdot \NTK^T)} = n  \cdot \cos^2(\NTK, \text{Id}_{H})\in [0, n]
\end{align}

\paragraph{Frobenius Norm}

The Frobenius norm of the NTK can be used to measure how well two operators align, e.g., through \(\|\NTK_1 - \NTK_2\|_F\). Comparing such operators can also be used to measure how much two learning rules, corresponding to modified operators, align~\citep{hazelden2026globalempiricalntkselfreferential}. The Frobenius norm of an NTK operator is simply \(\|\NTK\|_F^2 = \tr(\NTK \cdot \NTK^\top)\).

\section{Methods}

\label{sec:methods}

\subsection{The Hutch++ Trace Estimator} Throughout, we let \(G : H \rightarrow H\) denote a generic linear operator on the state-space \(H\). 
For any orthonormal basis \(Q = (q_i)_{i=1}^n\) of the state-space \(H\), the trace is \(\tr(G) := \sum_{i=1}^n q_i^T G q_i= \tr(Q^T G Q)\). This expression provides the simplest brute-force direct approach to computing the exact trace, via performing \(n\) matvec evaluations with \(G\). However, prior work has shown that far fewer matvecs can often be used to estimate \(\tr(G)\). A fast and simple approach for estimating this expression is Hutch++ \citep{meyer2021hutchoptimalstochastictrace}. Briefly, Hutch++ uses a randomized sketch of \(G\) to identify a low-dimensional subspace capturing much of the operator's action, then estimates the remaining trace contribution stochastically (see \citep{meyer2021hutchoptimalstochastictrace} and \citep{xtrace} for a detailed introduction). Crucially, modern analyses have shown that Hutch++ estimates \(\tr(G)\) for any \(G\) (not necessarily PSD) with variance \(O(1/m^2)\) as the sample count, \(m \rightarrow n\) \citep{xtrace}. 
Below is a summary of the Hutch++ algorithm for estimating an NTK trace. 

\begin{algorithm}[H]
\caption{Hutch++ NTK Trace Estimator}
\label{alg:hutchpp_ntk}
\begin{algorithmic}[1]
\Require Access to $\NTK(v)$ matvec for all \(v \in H\). Number of samples $m$
\Ensure Approximation to \(\tr(\NTK)\)
\State Draw $S, T \in \mathbb{R}^{n \times \frac{m}{3}}$ with i.i.d. \(\{+1, -1\}\) entries \hfill  \textcolor{gray}{// Or \(\mathcal{N}(0,1)\) entries}
\State $Q \gets \texttt{QR}(\texttt{NTK}(S))$ \hfill  \textcolor{gray}{// Basis for range of $\texttt{NTK}$}
\State $B \gets (I - QQ^\top) T$ \hfill  \textcolor{gray}{// Nullspace part of \(\NTK\) range}
\State \textbf{return} $\tr\left(\NTK(Q)^\top Q\right) + \frac{3}{m} \tr(\NTK(B)^T B)$  \hfill  \textcolor{gray}{// Low rank + residual estimates}
\end{algorithmic}
\end{algorithm}

\subsection{One-Sided NTK Trace Estimators (RHutch / FHutch)} As we show below, we find Hutch++ to be a robust and efficient estimator for traces of the NTK and related quantities. However, evaluating the NTK requires both \(\jvp\) and \(\vjp\) access (forward- and backward-mode AD). Sometimes one of these AD modes is not implemented or may bottleneck the trace estimation, as discussed below. To overcome this, we introduce two alternative Hutchinson-style estimators of the trace that rely on a single, specified AD direction (forward or backward) \citep{hutchinson1989stochastic}. Specifically, since the NTK has the form $JJ^\top$, the trace can be estimated as 
\(
\tr(JJ^\top)\;=\;\mathbb{E}_v\big[\|J^\top v\|_2^2\big]\;=\;\mathbb{E}_p\big[\|Jp\|_2^2\big]\),
where \(v \in H, p \in P\) are isotropic with entries from \(\mathcal{N}(0,1)\) in the state- or parameter-space, respectively. Thus, \(\tr(\NTK)\) can be estimated using only reverse-mode AD, ($G(v)=\vjp(v)$, \textit{RHutch}) or forward-mode AD, ($G(p)=\jvp(p)$, \textit{FHutch}). The algorithm below summarizes these estimators.

\begin{algorithm}[H]
\caption{One-Sided NTK Trace Estimators}
\label{alg:oneside_trace}
\begin{algorithmic}[1]
\Require Either define $G(v) = \vjp(v)$ for any \(v \in H\) and \(d = \dim(H)\), or $G(p) := \jvp(p)$ for any \(p \in P\) and \(d = \text{dim}(P)\). Number of samples $m$.
\Ensure Approximation to $\tr(\NTK)$
\State Draw $S, T \in \mathbb{R}^{d \times \frac{m}{2}}$ with i.i.d.\ $\{+1,-1\}$ entries \hfill \textcolor{gray}{// Or $\mathcal{N}(0,1)$}
\State $Q \gets \texttt{QR}(S)$ \hfill \textcolor{gray}{// Optional orthogonalization}
\State $B \gets (I - QQ^\top) T$ \hfill \textcolor{gray}{// Probes in the orthogonal complement}
\State \textbf{return} $\|G(Q)\|_F^2 + \frac{2}{m} \|G(B)\|_F^2$ \hfill  \textcolor{gray}{// Low rank + residual estimates}
\end{algorithmic}
\end{algorithm}

\vspace{-2.2em}

\paragraph{Viability} RHutch/FHutch sample different spaces and rely on a single AD mode, thus they can be beneficial when \(\dim P\) is much larger than \(\dim H\), or vice-versa. Also, for some architectures, GPU-accelerated implementations may only exist for one AD mode. The variance of both estimators scales as \(O(1/m)\) \citep{hutchinson1989stochastic}. For the higher \(O(1/m^2)\) accuracy, we recommend Hutch++, but show RHutch/FHutch can be faster in the small \(m\) regime, based on the architecture (see \sectn{numres}). Technically, the estimator in Algorithm~\ref{alg:oneside_trace} builds upon standard Hutchinson, instead using Haar-orthogonal probes and a residual control-variate, similar to Hutch++ but without sketching the NTK \citep{xtrace}. We used this form, since in practice it typically outperformed plain Hutchinson in our numerical experiments (details in Appendix~\ref{sec:aponeside}). 

\begin{table}[H]
\scriptsize
\centering
\begin{tabularx}{\linewidth}{>{\centering\arraybackslash}lYY}
\toprule
 & \textbf{Pros} & \textbf{Cons} \\
\midrule
\parbox[t]{3cm}{\centering \textbf{Hutch++}} 
& $O(1/m^2)$ variance guarantee and robust 
& May be bottlenecked by either forward- or backward-mode AD\\[12pt]
\parbox[t]{3cm}{\centering \textbf{One-Sided}} 
& Only either forward- or backward-AD 
& Worse $O(1/m)$ variance guarantee \\
\bottomrule
\end{tabularx}
\end{table}

\vspace{-2em}

\begin{figure}[t]
    \centering
    \includegraphics[width=.94\linewidth]{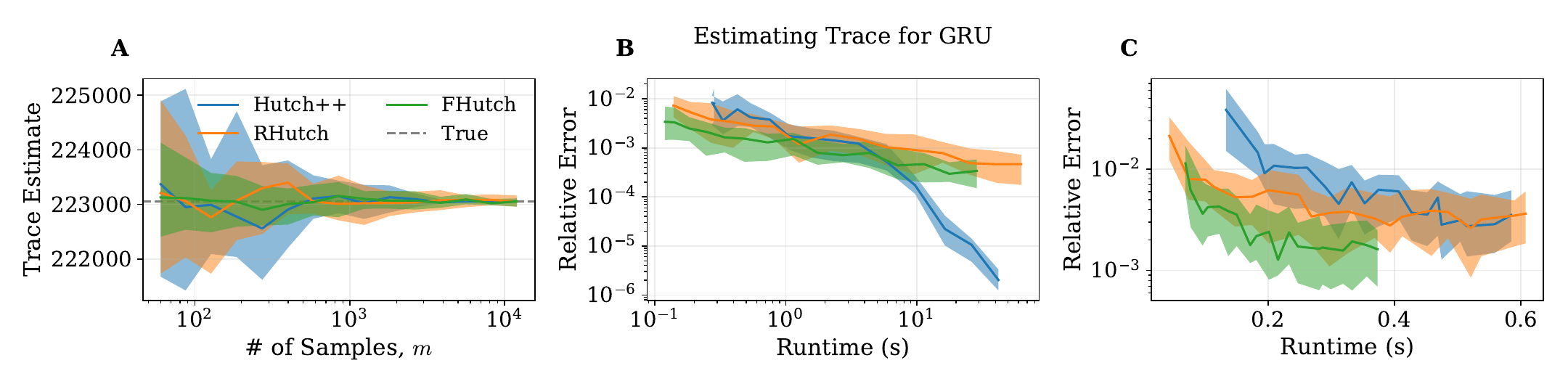}
    \caption{\textbf{Estimating \(\tr(\NTK)\) when \(h(\theta)\) tracks the hidden unit activations of a GRU neural network at every evaluation time.} \textbf{A} illustrates the trace estimates for Hutch++ and the one-sided Hutch estimator (Algorithms~\ref{alg:hutchpp_ntk} and~\ref{alg:oneside_trace}, see Figure~\ref{fig:mlp} for an analogous result for MLPs). \textbf{B} illustrates relative error \(\frac{|\tr(\NTK) - t_{m}|}{|\tr(\NTK)|}\) of each estimate, \(t_m\) versus runtime in seconds. In this case, exactly computing the trace with \(n\) matvecs took 1557.72 seconds (around 26 minutes). \textbf{C} shows the low-runtime region of B in more detail with linearly scaled runtime. In contrast to Fig.~\ref{fig:mlp}, FHutch performed fastest and both one-sided estimators outperformed Hutch++ for error above \(2 \cdot 10^{-3}\). Hutch++ was fastest for accuracy above 99.9\%.  }
    \label{fig:gru}
\end{figure}

\subsubsection{One-Sided Variants of NTK Statistics} For consistency and enabling future directions, in Appendix \sectn{aponeside} we prove a Proposition that facilitates estimation of NTK products of the form \(J_1 J_2^T J_3 J_4^T\) relying solely on either forward-mode or reverse-mode AD. We show that this enables one-sided estimation of the NTK norm, alignment, effective rank, and cross-trace in the style of Algorithm~\ref{alg:oneside_trace}. However, for the examples here, for which we are able to perform both forward-mode and reverse-mode AD, we find Hutch++ is simpler and generally outperforms these one-sided product estimators. Thus, the reported numerical experiments for these expressions below relied solely on Hutch++ \sectn{terms}. Variance reduction for these one-sided product estimators is an interesting direction for future work (discussed in \sectn{discuss}).

\section{Numerical Validation and Quantification of Speedup}

\label{sec:numres}

Here, we validate the estimators, measuring runtime versus accuracy for a GRU and a language model. Additionally, we validate on the classical output-space NTK of an MLP in Appendix~\ref{sec:apmlp}.

\subsection{NTK for the Hidden State of a Recurrent GRU}

\label{sec:gru}

\paragraph{Motivation} The NTK for recurrent models has been used to classify rich and lazy training regimes of recurrent models \citep{liu2024how}, characterize GD learning for linear RNNs \citep{bordelon2025dynamicallylearningintegraterecurrent}, and explain low-dimensional learning of latent dynamics \citep{pellegrino2023low, brandon2026mode, hazelden2026globalempiricalntkselfreferential}. However, the state dimension grows as \(n = B \cdot T \cdot H\) over batches, time, and hidden or output units, so explicitly forming the NTK requires an \(n\times n\) matrix. Faster matrix-free methods are therefore needed in the recurrent setting.

\paragraph{Setup} We let \(h(\theta)\) track the hidden state of a GRU recurrent network \citep{cho2014learning}. Similar to the MLP, we chose \(64\) hidden units, evaluated for \(T = 15\) timesteps on each input. Also, we provided the network with \(50\) random normal \(10\)-dimensional inputs at every timestep. Unlike the output of the MLP, tracking the hidden state here means \(h(\theta)\) is a 3-tensor in \(\R^{64 \times 15 \times 50}\), over \textit{all} timesteps, not just the final layer, so it has \(48,000\) entries.  By contrast, since the GRU re-uses the weights for each layer, the parameter count is \(14,592\). Hence, in this case \(\dim(P) < \dim(H)\).

\paragraph{Results} The results are summarized in Fig.~\ref{fig:gru}. For reference, the full trace computation was very slow, taking about \(26\) minutes, or \(1557.72\) seconds. By contrast, \(99\%\) accuracy could be obtained in less than \(0.1\) seconds with FHutch. An accuracy of \(99.99\%\) (relative error \(10^{-4}\)) took about 10 seconds with the Hutch++ estimator. Consistent with the prior findings, Hutch++ was fastest for high accuracy estimates, but the one-sided estimators could compete for low accuracy estimates. In contrast to the MLP output, however, here FHutch (using \(\jvp\)) was faster than Hutch++ for errors above around \(10^{-3}\). This is consistent with our expectation, since the number of entries in \(h(\theta)\) exceeds the parameter count in this case, and RHutch and FHutch sample from the state and parameter spaces, respectively. 

	\begin{figure}[t]
        \centering
        \includegraphics[width=.94\linewidth]{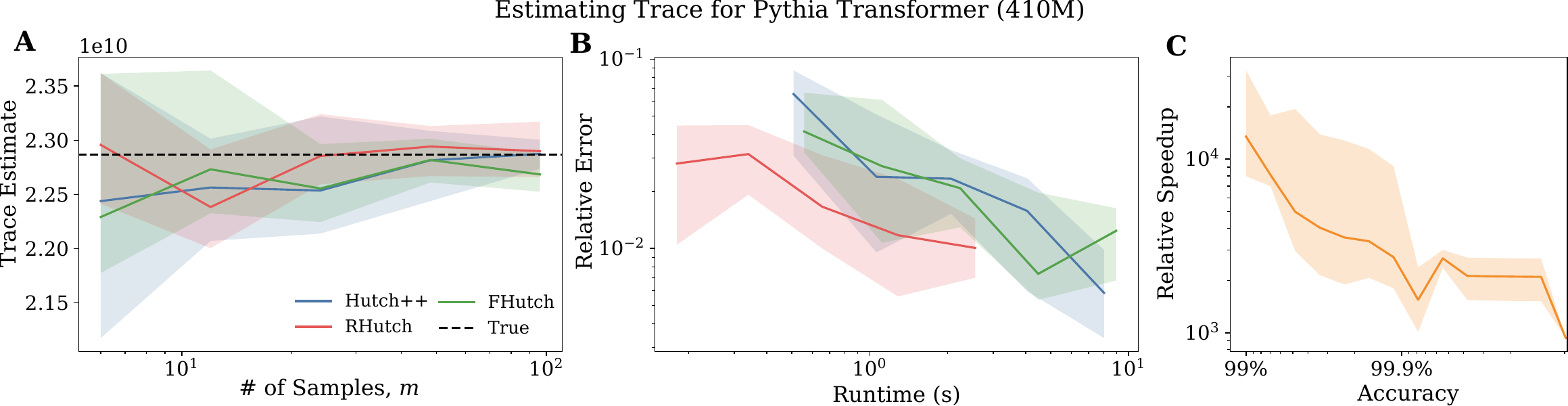}
        \caption{\textbf{Matrix-free estimation of \(\tr(\NTK)\) for the hidden-state representations of released Pythia Transformers.} The state \(h(\theta)\) contains 393,216 entries, and its Jacobian is taken with respect to 353,822,720 parameters. \textbf{A} shows trace estimates for Hutch++, RHutch, and FHutch against randomized sample count \(m\). \textbf{B} shows relative error \(\frac{|t_m-\tr(\NTK)|}{|\tr(\NTK)|}\) versus runtime. \textbf{C} shows the corresponding speedup (we plot the fastest of the 3 methods in the graph) relative to exact state-basis trace evaluation. Curves show medians over 20 estimates, with shaded regions indicating the \(25^{\mathrm{th}}\) and \(75^{\mathrm{th}}\) percentiles.
        \vspace{-0.7em}
        }
        \label{fig:transformer_trace_exact}
    \end{figure}

\subsection{NTK for the Hidden States of Transformer Language Models}
\label{sec:transformer}

\paragraph{Setup}
We next test whether the estimators remain practical for large Transformer state spaces. We use released Pythia-14M and Pythia-410M-deduped checkpoints, two GPT-NeoX language models from the Pythia suite \citep{biderman2023pythiasuiteanalyzinglarge}. Pythia provides publicly available models from the same family across a wide range of scales, making it a convenient controlled benchmark. For a fixed 16-token input, we define the state $h$ as the hidden representations of all tokens at the output of every Transformer block. This gives state dimensions $n=12{,}288$ and $n=393{,}216$ for Pythia-14M and Pythia-410M, respectively. Full architecture and implementation details are given in Appendix~\ref{app:pythia}.

\paragraph{Results}
Figure~\ref{fig:transformer_trace_exact} shows that the advantage of randomized trace estimation grows sharply with the Transformer state dimension. Exact state-basis evaluation takes 101.06 seconds for Pythia-14M and 8,328.58 seconds for Pythia-410M. At the 99\% accuracy target, the fastest randomized estimator is approximately $347$ times faster for Pythia-14M and $13{,}499$ times faster for Pythia-410M. For the larger model, percent-level estimates require only seconds: at $m=24$, RHutch reaches a median relative error of 1.66\% in 0.66 seconds, while at $m=96$, Hutch++ reaches 0.58\% median error in 8.03 seconds. Thus, hidden-state NTK statistics remain practical even when the state spans hundreds of thousands of coordinates and hundreds of millions of differentiated parameters.

\subsection{NTK Norm, Effective Rank and Alignment for MLP Trained on MNIST}

\label{sec:specials}

\begin{figure}[t]
    \centering
    \includegraphics[width=\linewidth]{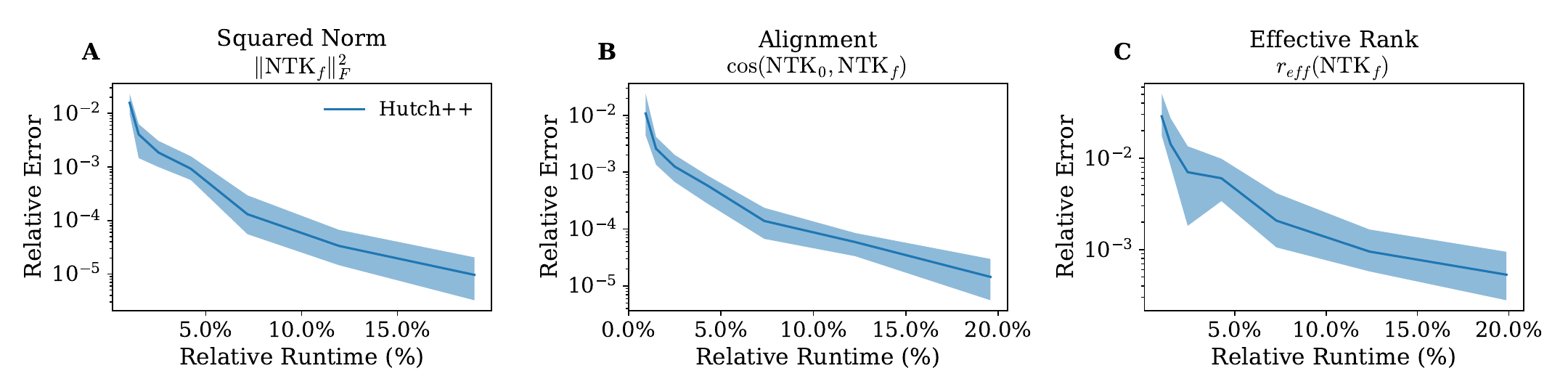}
    \caption{\textbf{Relative runtime versus error for an MLP trained on MNIST, estimating the Frobenius norm, kernel alignment, and the effective rank, respectively.} Here, \(\NTK_f\) denotes the NTK of the model post-training while \(\NTK_0\) is the NTK pre-training. \textbf{A--C} plot the relative error versus the exact value, computed with exact trace evaluation. We plot these versus the percentage of total runtime required to compute the exact expression in each case (range \(0\%\) to \(100\%\)). Curves correspond to the median over 50 re-evaluations with Hutch++ and the shaded regions illustrate the \(25^{\text{th}}\) and \(75^{\text{th}}\) percentiles.} 
    \label{fig:extras}
\end{figure}

\paragraph{Setup} We next apply the Hutch++ method to estimate the three trace-related quantities in Section~\ref{sec:terms}: Frobenius norm, alignment and effective rank. Specifically, we trained an MLP model on MNIST with a single hidden layer of 256 neurons and 10 output units, yielding about \(97\%\) test-set accuracy. We then evaluated the norm and effective rank of the NTK after training, \(\NTK_f\), and alignment between the NTK before and after training (\(\cos(\NTK_0, \NTK_f)\)). We used a testing set of 512 MNIST examples, yielding \(n = 5,120\). 

\paragraph{Results} We evaluated the three expressions with Hutch++ with results summarized in Fig.~\ref{fig:extras}, sweeping the sample count \(m\). There, we compare the estimation error to the relative runtime percentage (in the range 0-100), taken as the ratio of the median runtime over 50 re-runs of the estimator, divided by the time for evaluating the exact expression. In all three cases, we found Hutch++ attained 99\% accuracy (relative error \(10^{-2}\)) with \(m = 60\) samples and in about \(1\%\) of the total runtime for the exact trace calculation, i.e., a \(100\) times speedup. A relative error of \(10^{-4}\), i.e., \(99.99\%\) accuracy, was attained \(7.5\%\) of the total runtime for the norm and alignment and around \(25\%\) of the runtime for the effective rank. In total, we found that Hutch++ can attain high accuracies for all three expressions in a much shorter runtime. 

\section{Applications Enabled by Fast NTK Trace Estimation}
\label{sec:applications}

\subsection{Rich and Lazy Representation-Space Quantification}

\paragraph{Motivation} Rich and lazy learning differ in how strongly training changes a model's representation and NTK geometry \citep{jacot2018neural}. The lazy regime is interesting from a theoretical perspective because negligible NTK drift makes gradient flow tractable to analyze, yielding kernel gradient descent~\citep{chizat2019lazy}. The rich regime is interesting as it can lead to representations that map task structure enabling better generalization~\citep{farrell2023from, Flesch2022Orthogonal}. More recent work generalized these notions to recurrent networks and showed that there are previously unexplored features, such as the rank initial recurrent weights, \(\operatorname{rank}(W_0)\), modulate a transition from richer (low-rank) to lazier (high-rank) learning~\citep{liu2024how}. However, recent work has also emphasized that a simple rich-versus-lazy dichotomy can hide multiple phases of feature learning~\citep{chou2025featurelearninglazyrichdichotomy}.  Prior work~\citep{liu2024how} analyzing the hidden-state NTK in this context was limited by computation time, resulting in sparse measurements. Here, we show how fast trace estimation methods can extend these results to a denser sweep over training, more initialization ranks, and additional NTK statistics.

\begin{figure}[t]
    \centering
    \includegraphics[width=0.87\linewidth]{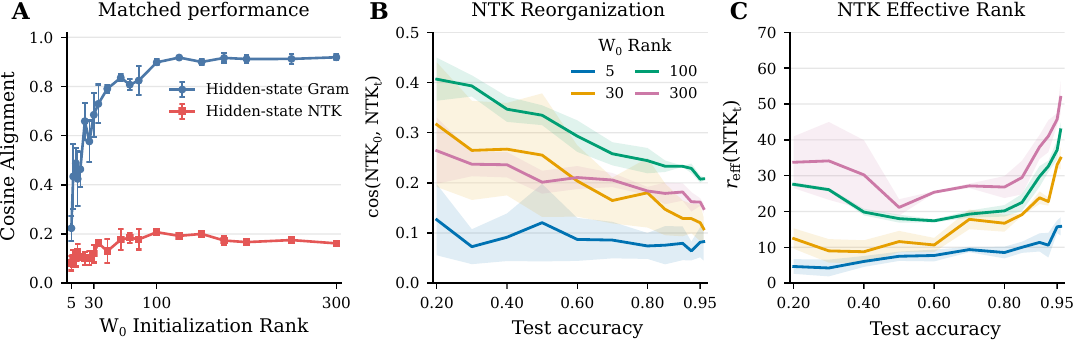}
    \caption{\textbf{Initialization-dependent rich and lazy reorganization of recurrent representations and NTK during learning on sequential MNIST.} As in~\citep{liu2024how}, we swept the rank of the recurrent weights \(W_0\) before training, with all models achieving above 95\% testing accuracy. \textbf{A} Quantifying drift (rich learning) in the RNN hidden state and hidden-state NTK before and after training per run. Each dot represents a distinct training run. \textbf{B} Hidden-state NTK alignment over training as test accuracy increases for four representative RNN initializations. \textbf{C} Corresponding NTK effective rank, showing continued expansion of tangent geometry during learning. Curves and error bars correspond to the standard error of the mean (SEM) over 3 seeds per run.}
\label{fig:helena}
\end{figure}

\paragraph{Setup}
We trained 300-unit RNNs on sequential MNIST while varying \(\operatorname{rank}(W_0)\) over 20 values and 3 random seeds per rank. Each model was trained with the same SGD setup as in~\citet{liu2024how}. Rather than comparing only initialization and the end of training, we evaluated the hidden-state NTK at 101 checkpoints per run, giving 6,060 checkpoint-level measurements. At each checkpoint, the same matrix-free estimator bundle provided NTK alignment and effective rank, with three estimator repeats for each reported value. The full trace-dependent statistics (involving \(54{,}900\) trace evaluations) were computed in 18.3 hours on a GTX Titan GPU. To separate geometric reorganization from differences in optimization speed, Figure~\ref{fig:helena} compares networks at matched levels of task performance rather than matched training time.

\paragraph{Results}
Figure~\ref{fig:helena}A recovers the broad rank-dependent rich-to-lazy trend reported in~\citet{liu2024how}, but at substantially higher resolution. Hidden-state representations become markedly more stable as initialization rank increases, while hidden-state NTK alignment shows a weaker dependence, with the strongest tangent reorganization concentrated at low rank. Panels~\textbf{B--C} show why repeated measurements are useful: as task performance improves, the NTK continues to drift from its initialization and generally expands in effective rank. A single initial-to-final alignment, therefore, captures the overall rich-to-lazy trend but misses structure in the trajectory of tangent geometry.

\subsection{Teacher NTK Alignment as a Data-Scarce Regularizer}
\label{sec:fashion_mnist}

\paragraph{Motivation}
The matrix-free construction developed above makes it inexpensive to compare tangent geometry during training.
In particular, the runtime results for MLPs in Appendix~\ref{sec:apmlp} suggest that stochastic estimates of operator alignment can be incorporated directly into optimization at modest additional cost. As a second motivating example, we examine using the NTK as a regularizer during GD training, specifically as a source of additional geometric information in a data-limited knowledge distillation setting. 
This is motivated by recent work of~\citet{xiang2025neuraltangentknowledgedistillation}, which used the output-space NTK as a regularizer for improving training of an optical neural network but did not examine generalization and was limited computationally, relying on estimating the NTK matrix (as in~\citet{wang2023ntksapimprovingneuralnetwork}) to compute its statistics.

\paragraph{Setup}
We first train a teacher network on the full Fashion-MNIST training set.  This is a single-hidden-layer MLP with 512 hidden units. We then independently initialize student networks of the same architecture using only $N\in\{250,500,1000\}$ training examples. Students minimize $\mathcal{L}=(1-\lambda)\mathcal{L}_{\mathrm{CE}}+\lambda\left(1-\cos(\NTK_S,\NTK_T)\right)$, where $\NTK_S$ and $\NTK_T$ denote student and teacher NTK operators evaluated on the student training data. Alignment is estimated entirely matrix-free using Hutch++ with 12 probes. We consider both output-logit and hidden-state NTKs, and compare against otherwise identical students trained with cross-entropy alone ($\lambda=0$). All students reach essentially $100\%$ training accuracy, allowing us to isolate differences in generalization among interpolating solutions.

\begin{figure}[t]
    \centering
    \includegraphics[width=0.93\textwidth]{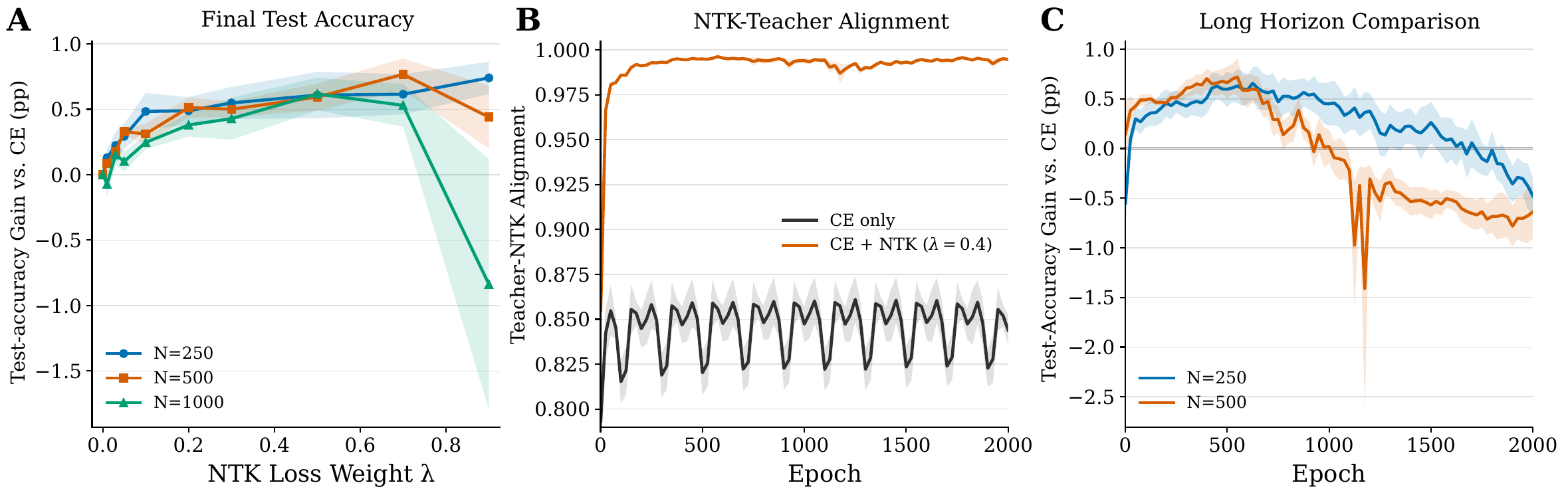}
    \caption{\textbf{Teacher-student NTK alignment as a regularizer for data-limited Fashion-MNIST students.} A student is trained on a pretrained teacher's output with limited training data and an NTK regularization term with factor \(\lambda\). \textbf{A} Final test-accuracy gain relative to matched cross-entropy-only students as a function for training set size \(N\in\{250,500,1000\}\). Positive values indicate improved generalization  \textbf{B} Teacher-NTK alignment during training for cross-entropy-only and NTK-regularized students (\(\lambda=0.4\)), showing rapid and sustained alignment with the teacher geometry. \textbf{C} Test-accuracy gain relative to cross-entropy-only training over a longer training horizon for \(N=250\) and \(N=500\). Shaded regions in A and C correspond to SEM over reruns.}
    \label{fig:fashion_ntk}
\end{figure}

\paragraph{Results} The Table below and Figure~\ref{fig:fashion_ntk} summarize the results. NTK matching yields modest generalization gains, strongest in the very data-scarce setting, suggesting that it could be used in future directions to impart additional geometrical structure in knowledge distillation.

\vspace{-1.0em}
\begin{table}[H]
    \centering
    \scriptsize
    \begin{tabular}{llrrr}
        \toprule
        NTK Space & $\lambda$ & $N$ &
        $\Delta$ test acc. (pp) & Paired wins \\
        \midrule
        Logits & 0.7 & 250  & $+0.62 \pm 0.15$ & $5/6$ \\
        Logits & 0.7 & 500  & $+0.77 \pm 0.12$ & $6/6$ \\
        Logits & 0.7 & 1000 & $+0.53 \pm 0.16$ & $6/6$ \\
        Hidden & 0.5 & 250  & $+0.98 \pm 0.22$ & $3/3$ \\
        Hidden & 0.5 & 500  & $+0.49 \pm 0.23$ & $3/3$ \\
        Hidden & 0.5 & 1000 & $-0.21 \pm 0.16$ & $1/3$ \\
        \bottomrule
    \end{tabular}
   \label{tab:fashion_ntk}
\end{table}
\vspace{-2.5em}

\section{Discussion}
\label{sec:discuss}

Finite-width NTK analysis can be prohibitively expensive, especially for state-space NTKs spanning batches, hidden units, timesteps, layers, or tokens. We show that useful quantities, including the trace, Frobenius norm, alignment, and effective rank, can instead be estimated from matrix-free NTK products using randomized trace estimation. We also introduce one-sided estimators using only forward- or reverse-mode AD, which can be advantageous at coarse tolerances or when one AD direction is cheaper. Across MLPs, recurrent networks, and Transformers, these methods make finite-width NTK statistics practical at scales where direct evaluation is expensive.

This enables repeated NTK measurements during training rather than only isolated endpoint calculations. We use this to track tangent-geometry changes during rich versus lazy recurrent learning and to include teacher-student NTK alignment directly in a training objective, where it modestly improves generalization in data-scarce settings.

\paragraph{Limitations and Future Directions}
The one-sided estimators do not exploit the NTK's low-rank output space as effectively as Hutch++, so their advantage is largely confined to coarse accuracy regimes. Output-informed one-sided estimators are a natural future direction. Partial trace estimation is another promising extension, allowing tensor-valued NTKs to be reduced while retaining temporal, spatial, or feature structure~\citep{Chen_2024,XuZhu2024_UniformConcentrationNTK,hazelden2026globalempiricalntkselfreferential}. Another related direction is to use finer grain metrics. For example, since the state-space can be empirically selected here, one can consider the NTK on a specific, restricted set of task inputs. Accumulating and comparing the statistics of these NTKs can build a finer grain model of how the NTK depends on task structure, for example, in a multi-task setting where notions of re-use and compositional hidden state structure are of interest~\citep{turner2023simplicity, driscoll2024flexible}.  

\paragraph{\textbf{Code Availability}} The NTK and related operators are implemented in a matrix-free way in our package \url{https://anonymous.4open.science/r/gentk-47B8}. Timing figures can be reproduced by running \verb|experiments/trace_paper/main.py|. 

\paragraph{LLM Usage Statement}
Large language models were used for assistance with experimental code (specifically, the two applications in Section 6) and manuscript editing. The methodology, experimental design, core implementation, testing, and validation were developed by the authors without LLMs.

\bibliography{refs}

@article{Flesch2022Orthogonal,
  author    = {Flesch, Timo and Juechems, Keno and Dumbalska, Tsvetomira and Saxe, Andrew and Summerfield, Christopher},
  title     = {Orthogonal representations for robust context-dependent task performance in brains and neural networks},
  journal   = {Neuron},
  year      = {2022},
  volume    = {110},
  number    = {7},
  pages     = {1258--1270.e11},
  doi       = {10.1016/j.neuron.2022.01.005},
  publisher = {Elsevier},
  url       = {https://doi.org/10.1016/j.neuron.2022.01.005}
}

@misc{black2022gptneox20bopensourceautoregressivelanguage,
      title={GPT-NeoX-20B: An Open-Source Autoregressive Language Model}, 
      author={Sid Black and Stella Biderman and Eric Hallahan and Quentin Anthony and Leo Gao and Laurence Golding and Horace He and Connor Leahy and Kyle McDonell and Jason Phang and Michael Pieler and USVSN Sai Prashanth and Shivanshu Purohit and Laria Reynolds and Jonathan Tow and Ben Wang and Samuel Weinbach},
      year={2022},
      eprint={2204.06745},
      archivePrefix={arXiv},
      primaryClass={cs.CL},
      url={https://arxiv.org/abs/2204.06745}, 
}

@misc{biderman2023pythiasuiteanalyzinglarge,
      title={Pythia: A Suite for Analyzing Large Language Models Across Training and Scaling}, 
      author={Stella Biderman and Hailey Schoelkopf and Quentin Anthony and Herbie Bradley and Kyle O'Brien and Eric Hallahan and Mohammad Aflah Khan and Shivanshu Purohit and USVSN Sai Prashanth and Edward Raff and Aviya Skowron and Lintang Sutawika and Oskar van der Wal},
      year={2023},
      eprint={2304.01373},
      archivePrefix={arXiv},
      primaryClass={cs.CL},
      url={https://arxiv.org/abs/2304.01373}, 
}

@article{kramer1985stabilization,
  title = {Stabilization of unstable states, relaxation, and critical slowing down in a bistable system},
  author = {Kramer, Jesse and Ross, John},
  journal = {The Journal of Chemical Physics},
  volume = {83},
  number = {12},
  pages = {6234--6241},
  year = {1985},
  publisher = {American Institute of Physics},
}

@misc{hazelden2026globalempiricalntkselfreferential,
  title = {The Global Empirical NTK: Self-Referential Bias and Dimensionality of Gradient Descent Learning},
  author = {James Hazelden and Laura Driscoll and Eli Shlizerman and Eric Shea-Brown},
  year = {2026},
  eprint = {2605.08746},
  archiveprefix = {arXiv},
  primaryclass = {cs.LG},
  url = {https://arxiv.org/abs/2605.08746},
}

@article{anderson2023sloppy,
  title = {Sloppy Model Analysis Identifies Bifurcation Parameters without Normal Form Analysis},
  author = {Anderson, Christian N. K. and Transtrum, Mark K.},
  journal = {Physical Review E},
  volume = {108},
  number = {6},
  pages = {064215},
  year = {2023},
  doi = {10.1103/PhysRevE.108.064215},
}

@inproceedings{pellegrino2023low,
  title = {Low Tensor Rank Learning of Neural Dynamics},
  author = {Arthur Pellegrino and N. Alex Cayco Gajic and Angus Chadwick},
  booktitle = {Advances in Neural Information Processing Systems},
  volume = {36},
  pages = {11674--11702},
  year = {2023},
}

@article{yang2020tensorprogramsII,
  title = {Tensor Programs II: Neural Tangent Kernel for Any Architecture},
  author = {Greg Yang},
  journal = {arXiv preprint arXiv:2006.14548},
  year = {2020},
}

@article{feng2023deqntk,
  title = {On the Neural Tangent Kernel of Equilibrium Models},
  author = {Zhili Feng and J. Zico Kolter},
  journal = {arXiv preprint arXiv:2310.14062},
  year = {2023},
}

@inproceedings{mohamadi2023fast,
  title = {A Fast, Well-Founded Approximation to the Empirical Neural Tangent Kernel},
  author = {Mohamad Amin Mohamadi and Wonho Bae and Danica J. Sutherland},
  booktitle = {Proceedings of the 40th International Conference on Machine Learning},
  series = {Proceedings of Machine Learning Research},
  volume = {202},
  pages = {25011--25035},
  year = {2023},
}

@misc{xiang2025neuraltangentknowledgedistillation,
  title = {Neural Tangent Knowledge Distillation for Optical Convolutional Networks},
  author = {Jinlin Xiang and Minho Choi and Yubo Zhang and Zhihao Zhou and Arka Majumdar and Eli Shlizerman},
  year = {2025},
  eprint = {2508.08421},
  archiveprefix = {arXiv},
  primaryclass = {cs.CV},
  url = {https://arxiv.org/abs/2508.08421},
}

@misc{shan2022theoryneuraltangentkernel,
  title = {A Theory of Neural Tangent Kernel Alignment and Its Influence on Training},
  author = {Haozhe Shan and Blake Bordelon},
  year = {2022},
  eprint = {2105.14301},
  archiveprefix = {arXiv},
  primaryclass = {stat.ML},
  url = {https://arxiv.org/abs/2105.14301},
}

@inproceedings{meyer2021hutchoptimalstochastictrace,
  title = {Hutch++: Optimal Stochastic Trace Estimation},
  author = {Meyer, Raphael A. and Musco, Cameron and Musco, Christopher and Woodruff, David P.},
  booktitle = {Symposium on Simplicity in Algorithms (SOSA)},
  pages = {142--155},
  year = {2021},
  publisher = {SIAM},
  doi = {10.1137/1.9781611976496.16},
  eprint = {2010.09649},
  archiveprefix = {arXiv},
  url = {https://arxiv.org/abs/2010.09649},
}

@inproceedings{bordelon2020spectrum,
  title = {Spectrum Dependent Learning Curves in Kernel Regression and Wide Neural Networks},
  author = {Bordelon, Blake and Canatar, Abdulkadir and Pehlevan, Cengiz},
  booktitle = {International Conference on Machine Learning},
  pages = {1024--1034},
  year = {2020},
  publisher = {PMLR},
}

@misc{novak2019neuraltangentsfasteasy,
  title = {Neural Tangents: Fast and Easy Infinite Neural Networks in Python},
  author = {Roman Novak and Lechao Xiao and Jiri Hron and Jaehoon Lee and Alexander A. Alemi and Jascha Sohl-Dickstein and Samuel S. Schoenholz},
  year = {2019},
  eprint = {1912.02803},
  archiveprefix = {arXiv},
  primaryclass = {stat.ML},
  url = {https://arxiv.org/abs/1912.02803},
}

@inproceedings{novak2022fastfinitewidthneural,
  title = {Fast Finite Width Neural Tangent Kernel},
  author = {Novak, Roman and Sohl-Dickstein, Jascha and Schoenholz, Samuel S.},
  booktitle = {Proceedings of the 39th International Conference on Machine Learning},
  pages = {17018--17044},
  year = {2022},
  editor = {Chaudhuri, Kamalika and Jegelka, Stefanie and Song, Le and Szepesv{\'a}ri, Csaba and Niu, Gang and Sabato, Sivan},
  volume = {162},
  series = {Proceedings of Machine Learning Research},
  month = {17--23 Jul},
  publisher = {PMLR},
  url = {https://proceedings.mlr.press/v162/novak22a.html},
}

@inproceedings{pmlr-v139-yang21c,
  title = {Tensor Programs IV: Feature Learning in Infinite-Width Neural Networks},
  author = {Yang, Greg and Hu, Edward J.},
  booktitle = {Proceedings of the 38th International Conference on Machine Learning},
  pages = {11727--11737},
  year = {2021},
  editor = {Meila, Marina and Zhang, Tong},
  volume = {139},
  series = {Proceedings of Machine Learning Research},
  month = {18--24 Jul},
  publisher = {PMLR},
  url = {https://proceedings.mlr.press/v139/yang21c.html},
}

@article{farrell2023from,
  title = {From lazy to rich to exclusive task representations in neural networks and neural codes},
  author = {Farrell, M. and Recanatesi, S. and Shea-Brown, E.},
  journal = {Current Opinion in Neurobiology},
  volume = {83},
  pages = {102780},
  year = {2023},
  doi = {10.1016/j.conb.2023.102780},
  note = {Epub 2023 Sep 25},
  pmid = {37757585},
}

@article{driscoll2024flexible,
  title = {Flexible multitask computation in recurrent networks utilizes shared dynamical motifs},
  author = {Driscoll, Laura N. and Shenoy, Krishna V. and Sussillo, David},
  journal = {Nature Neuroscience},
  volume = {27},
  pages = {1349--1363},
  year = {2024},
  doi = {10.1038/s41593-024-01668-6},
}

@inproceedings{turner2023simplicity,
  author = {Turner, Elia and Barak, Omri},
  booktitle = {Advances in Neural Information Processing Systems},
  editor = {A. Oh and T. Naumann and A. Globerson and K. Saenko and M. Hardt and S. Levine},
  pages = {25495--25507},
  publisher = {Curran Associates, Inc.},
  title = {The Simplicity Bias in Multi-Task RNNs: Shared Attractors, Reuse of Dynamics, and Geometric Representation},
  url = {https://proceedings.neurips.cc/paper_files/paper/2023/file/50d6dbc809b0dc96f7f1090810537acc-Paper-Conference.pdf},
  volume = {36},
  year = {2023},
}

@inproceedings{chizat2019lazy,
  title = {On Lazy Training in Differentiable Programming},
  author = {Chizat, L{\'e}na{\"i}c and Oyallon, Edouard and Bach, Francis},
  booktitle = {Advances in Neural Information Processing Systems},
  volume = {32},
  year = {2019},
}

@inproceedings{liu2024how,
  title = {How connectivity structure shapes rich and lazy learning in neural circuits},
  author = {Yuhan Helena Liu and Aristide Baratin and Jonathan Cornford and Stefan Mihalas and Eric Todd Shea-Brown and Guillaume Lajoie},
  booktitle = {The Twelfth International Conference on Learning Representations},
  year = {2024},
  url = {https://openreview.net/forum?id=slSmYGc8ee},
}

@article{alemohammad2020recurrent,
  title = {The Recurrent Neural Tangent Kernel},
  author = {Alemohammad, Saman and Wang, Zhuang and Balestriero, Remi and Baraniuk, Richard},
  journal = {Advances in Neural Information Processing Systems},
  volume = {33},
  year = {2020},
}

@article{mei2018mean,
  title = {Mean-field theory of two-layer neural networks: dimension-free bounds and kernel limit},
  author = {Mei, Song and Misiakiewicz, Tomasz and Montanari, Andrea},
  journal = {Advances in Neural Information Processing Systems},
  volume = {31},
  year = {2018},
}

@article{jacot2018neural,
  title = {Neural Tangent Kernel: Convergence and Generalization in Neural Networks},
  author = {Jacot, Arthur and Gabriel, Franck and Hongler, Cl{\'e}ment},
  journal = {Advances in Neural Information Processing Systems},
  volume = {31},
  year = {2018},
}

@misc{paszke2019pytorchimperativestylehighperformance,
  title = {PyTorch: An Imperative Style, High-Performance Deep Learning Library},
  author = {Adam Paszke and Sam Gross and Francisco Massa and Adam Lerer and James Bradbury and Gregory Chanan and Trevor Killeen and Zeming Lin and Natalia Gimelshein and Luca Antiga and Alban Desmaison and Andreas K{\"o}pf and Edward Yang and Zach DeVito and Martin Raison and Alykhan Tejani and Sasank Chilamkurthy and Benoit Steiner and Lu Fang and Junjie Bai and Soumith Chintala},
  year = {2019},
  eprint = {1912.01703},
  archiveprefix = {arXiv},
  primaryclass = {cs.LG},
  url = {https://arxiv.org/abs/1912.01703},
}

@misc{bordelon2025dynamicallylearningintegraterecurrent,
  title = {Dynamically Learning to Integrate in Recurrent Neural Networks},
  author = {Blake Bordelon and Jordan Cotler and Cengiz Pehlevan and Jacob A. Zavatone-Veth},
  year = {2025},
  eprint = {2503.18754},
  archiveprefix = {arXiv},
  primaryclass = {q-bio.NC},
  url = {https://arxiv.org/abs/2503.18754},
}

@article{hutchinson1989stochastic,
  author = {Hutchinson, M. F.},
  title = {A Stochastic Estimator of the Trace of the Influence Matrix for Laplacian Smoothing Splines},
  journal = {Communications in Statistics - Simulation and Computation},
  year = {1989},
  volume = {18},
  number = {3},
  pages = {1059--1076},
  doi = {10.1080/03610918908812806},
}

@misc{atanasov2021neuralnetworkskernellearners,
  title = {Neural Networks as Kernel Learners: The Silent Alignment Effect},
  author = {Alexander Atanasov and Blake Bordelon and Cengiz Pehlevan},
  year = {2021},
  eprint = {2111.00034},
  archiveprefix = {arXiv},
  primaryclass = {stat.ML},
  url = {https://arxiv.org/abs/2111.00034},
}

@inproceedings{arora2019exact,
  title = {On Exact Computation with an Infinitely Wide Neural Net},
  author = {Arora, Sanjeev and Du, Simon S. and Hu, Wei and Li, Zhiyuan
            and Salakhutdinov, Ruslan and Wang, Ruosong},
  booktitle = {Advances in Neural Information Processing Systems},
  volume = {32},
  year = {2019},
  url = {https://arxiv.org/abs/1904.11955},
}

@article{martinsson2020randomized,
  title = {Randomized Numerical Linear Algebra: Foundations and Algorithms},
  author = {Martinsson, Per-Gunnar and Tropp, Joel A.},
  journal = {Acta Numerica},
  volume = {29},
  pages = {403--572},
  year = {2020},
  doi = {10.1017/S0962492920000021},
}

@article{boulle2024operator,
  title = {Operator Learning without the Adjoint},
  author = {Boull{\'e}, Nicolas and Halikias, Diana and Otto, Samuel E.
            and Townsend, Alex},
  journal = {Journal of Machine Learning Research},
  volume = {25},
  number = {364},
  pages = {1--54},
  year = {2024},
  url = {https://www.jmlr.org/papers/v25/24-0162.html},
}

@misc{chou2025featurelearninglazyrichdichotomy,
      title={Feature Learning beyond the Lazy-Rich Dichotomy: Insights from Representational Geometry}, 
      author={Chi-Ning Chou and Hang Le and Yichen Wang and SueYeon Chung},
      year={2025},
      eprint={2503.18114},
      archivePrefix={arXiv},
      primaryClass={cs.LG},
      url={https://arxiv.org/abs/2503.18114}, 
}

@article{persson2022improved,
  title = {Improved Variants of the {Hutch++} Algorithm for Trace Estimation},
  author = {Persson, David and Cortinovis, Alice and Kressner, Daniel},
  journal = {SIAM Journal on Matrix Analysis and Applications},
  volume = {43},
  number = {3},
  pages = {1162--1185},
  year = {2022},
  doi = {10.1137/21M1447623},
}

@misc{chen2026variancesensitivitybaseddimensionsbrain,
      title={Beyond variance: sensitivity-based dimensions in brain networks underlie individual differences in cognitive ability}, 
      author={Sida Chen and Siqi Yang and Zhao Chang and Taro Toyoizumi and Werner Sommer and Lianchun Yu and Qian-Yuan Tang and Changsong Zhou},
      year={2026},
      eprint={2501.19106},
      archivePrefix={arXiv},
      primaryClass={q-bio.NC},
      url={https://arxiv.org/abs/2501.19106}, 
}

@article{zvonek2025conthutch,
  title = {{ContHutch++}: Stochastic Trace Estimation for Implicit Integral Operators},
  author = {Zvonek, Jennifer and Horning, Andrew J. and Townsend, Alex},
  journal = {SIAM Journal on Numerical Analysis},
  volume = {63},
  number = {1},
  pages = {334--359},
  year = {2025},
  doi = {10.1137/23M1614365},
}

@inproceedings{lee2019wide,
  title={Wide Neural Networks of Any Depth Evolve as Linear Models Under Gradient Descent},
  author={Lee, Jaehoon and Xiao, Lechao and Schoenholz, Samuel S. and Bahri, Yasaman
          and Novak, Roman and Sohl-Dickstein, Jascha and Pennington, Jeffrey},
  booktitle={Advances in Neural Information Processing Systems},
  year={2019}
}

@misc{dangel2025positioncurvaturematricesdemocratized,
      title={Position: Curvature Matrices Should Be Democratized via Linear Operators}, 
      author={Felix Dangel and Runa Eschenhagen and Weronika Ormaniec and Andres Fernandez and Lukas Tatzel and Agustinus Kristiadi},
      year={2025},
      eprint={2501.19183},
      archivePrefix={arXiv},
      primaryClass={cs.LG},
      url={https://arxiv.org/abs/2501.19183}, 
}

@inproceedings{
brandon2026mode,
title={Mode Collapse Emerges from Low-Rank Biases in the Learning Dynamics of Generative Models},
author={Julian Brandon and Bruno Loureiro and N Alex Cayco Gajic and Arthur Pellegrino},
booktitle={High-dimensional Learning Dynamics 2026},
year={2026},
url={https://openreview.net/forum?id=sp206XLwHr}
}

@misc{kumar2024grokkingtransitionlazyrich,
  title = {Grokking as the Transition from Lazy to Rich Training Dynamics},
  author = {Tanishq Kumar and Blake Bordelon and Samuel J. Gershman and Cengiz Pehlevan},
  year = {2024},
  eprint = {2310.06110},
  archiveprefix = {arXiv},
  primaryclass = {stat.ML},
  url = {https://arxiv.org/abs/2310.06110},
}

@article{internalDynamicsNTK2025,
  title   = {Beyond Scaling Curves: Internal Dynamics of Neural Networks Through the NTK Lens},
  author  = {Nikolaou, Konstantin and Krippendorf, Sven and Tovey, Samuel and Holm, Christian},
  year    = {2025},
  journal = {arXiv preprint arXiv:2507.05035},
  url     = {https://arxiv.org/abs/2507.05035}
}

@inproceedings{murray2023characterizing,
  title = {Characterizing the Spectrum of the NTK via a Power Series Expansion},
  author = {Murray, Michael and Jin, Hui and Bowman, Benjamin and Mont\'ufar, Guido},
  booktitle = {International Conference on Learning Representations (ICLR) 2023},
  year = {2023},
  note = {Poster},
  url = {https://openreview.net/forum?id=Tvms8xrZHyR},
}

@misc{wang2023ntksapimprovingneuralnetwork,
  title = {NTK-SAP: Improving neural network pruning by aligning training dynamics},
  author = {Yite Wang and Dawei Li and Ruoyu Sun},
  year = {2023},
  eprint = {2304.02840},
  archiveprefix = {arXiv},
  primaryclass = {cs.LG},
  url = {https://arxiv.org/abs/2304.02840},
}

@article{xtrace,
  author = {Epperly, Ethan N. and Tropp, Joel A. and Webber, Robert J.},
  title = {XTrace: Making the Most of Every Sample in Stochastic Trace Estimation},
  journal = {SIAM Journal on Matrix Analysis and Applications},
  volume = {45},
  number = {1},
  pages = {1--23},
  year = {2024},
  doi = {10.1137/23M1548323},
  url = {https://doi.org/10.1137/23M1548323},
}

@article{XuZhu2024_UniformConcentrationNTK,
  title = {Overparametrized Multi-layer Neural Networks: Uniform Concentration of Neural Tangent Kernel and Convergence of Stochastic Gradient Descent},
  author = {Jiaming Xu and Hanjing Zhu},
  journal = {Journal of Machine Learning Research},
  volume = {25},
  number = {1},
  pages = {1--83},
  year = {2024},
  url = {https://jmlr.org/papers/v25/23-0740.html},
}

@article{Chen_2024,
  title = {Faster Randomized Partial Trace Estimation},
  volume = {46},
  issn = {1095-7197},
  url = {http://dx.doi.org/10.1137/23M1620399},
  doi = {10.1137/23M1620399},
  number = {6},
  journal = {SIAM Journal on Scientific Computing},
  publisher = {Society for Industrial and Applied Mathematics},
  author = {Chen, Tyler and Chen, Robert and Li, Kevin and Nzeuton, Skai and Pan, Yilu and Wang, Yixin},
  year = {2024},
  month = nov,
  pages = {A3427--A3447},
}

@misc{zandieh2021scalingneuraltangentkernels,
  title = {Scaling Neural Tangent Kernels via Sketching and Random Features},
  author = {Amir Zandieh and Insu Han and Haim Avron and Neta Shoham and Chaewon Kim and Jinwoo Shin},
  year = {2021},
  eprint = {2106.07880},
  archiveprefix = {arXiv},
  primaryclass = {cs.LG},
  url = {https://arxiv.org/abs/2106.07880},
}

@inproceedings{sharma2025relationship,
  title = {On the Relationship Between Neural Tangent Kernel Frobenius Distance and Distillation Sample Complexity},
  author = {Sharma, Arnav and Wez, Ahmed and Srikumar, Karthik},
  booktitle = {Workshop on Lock-LLM at Neural Information Processing Systems (NeurIPS 2025)},
  year = {2025},
  url = {https://openreview.net/pdf/89e812786af493b2305487f7a6a1689d41b324ec.pdf},
  note = {Poster},
}

@article{DBLP:journals/corr/abs-1906-08034,
  author = {Mario Geiger and Stefano Spigler and Arthur Jacot and Matthieu Wyart},
  title = {Disentangling feature and lazy learning in deep neural networks: an empirical study},
  journal = {CoRR},
  volume = {abs/1906.08034},
  year = {2019},
  url = {http://arxiv.org/abs/1906.08034},
  eprinttype = {arXiv},
  eprint = {1906.08034},
}

@article{cho2014learning,
  title = {Learning Phrase Representations using RNN Encoder--Decoder for Statistical Machine Translation},
  author = {Cho, Kyunghyun and Van Merrienboer, Bart and Gulcehre, Caglar and Bahdanau, Dzmitry and Bougares, Fethi and Schwenk, Holger and Bengio, Yoshua},
  journal = {arXiv preprint arXiv:1406.1078},
  year = {2014},
}
\bibliographystyle{iclr2027_conference}

\newpage
\clearpage
\appendix

\section*{Appendix}
\label{sec:app}

\section{Implementation Details}
All numerical experiments were implemented in PyTorch \citep{paszke2019pytorchimperativestylehighperformance}. Unless stated otherwise, the smaller MLP and recurrent validation experiments were run on a CPU. The Pythia experiments were run on an NVIDIA L40S GPU, as described in \sectn{transformer}. We compute the exact trace by applying the NTK to the \(n\) standard basis vectors \(e_i\in H\), without explicitly storing the full NTK. Where batching is used, these products are accelerated with \texttt{torch.vmap}. As a runtime sanity check, Hutch++ with \(m=n\) matvec evaluations typically took \(80\)--\(120\%\) of the direct estimator runtime, giving a speedup close to \(1\times\). This suggests that the reported speedups are not artifacts of substantially different implementations.

For the Pythia experiment, products were vectorized with chunks of size 8 on the NVIDIA L40S GPU.  We compared Hutch++, RHutch, and FHutch with $m\in\{6,12,24,48,96\}$ using 20 independent repeats for each case.

\subsection{Hutch++ Implementation Details}

The allocation of the total probe budget between the range sketch and residual estimator is a tunable hyperparameter. Algorithm~\ref{alg:hutchpp_ntk} uses the simple equal allocation \(q=r=m/3\), so the total number of NTK products is \(2q+r=m\). The numerical experiments use the experiment-specific allocations stated below; these do not change the Hutch++ decomposition.

\section{One-Sided NTK Product Estimator} 
\label{sec:aponeside}
In the main text, we show in Algorithm~\ref{alg:oneside_trace} that the NTK trace can be approximated with a Hutchinson estimator using only one AD mode (reverse, \(\vjp\), or forward, \(\jvp\)). Here, we show this extends to products of NTKs, yielding one-sided estimators for the norm, alignment, effective rank, as well as the case where \(\NTK = J_1 J_2^T\) with distinct Jacobians on the left and right side (e.g., evaluated on the testing and training set, respectively). This result is summarized in the following Proposition. 

\begin{proposition}
\label{prop:est}
Let $J_1, J_2, J_3, J_4 : P \rightarrow S$ be operators mapping the parameter space to the state space. Then, letting $u, v \in H$ be independent random vectors such that 
$\mathbb{E}[u] = \mathbb{E}[v] = 0$ and 
$\mathbb{E}[uu^\top] = \mathbb{E}[vv^\top] = I$ (e.g., uniform \(\{+1, -1\}\) or \(\mathcal{N}(0,1)\) entries). Then,
\[
\operatorname{tr}\!\left(J_1 J_2^\top J_4 J_3^\top\right)
= 
\mathbb{E}\!\left[
\left\langle J_1^\top u,\, J_2^\top v \right\rangle \cdot 
\left\langle J_4^\top u,\, J_3^\top v \right\rangle
\right],
\]
yielding a reverse-mode only estimator. Likewise, probing in the parameter space: if \(p, q \in P\) denote random vectors with $\mathbb{E}[p] = \mathbb{E}[q] = 0$, $\mathbb{E}[p p^\top] = \mathbb{E}[q q^\top] = I$, then 
\[
\operatorname{tr}\!\left(J_1 J_2^\top J_3 J_4^\top\right)
= 
\mathbb{E}\!\left[
\left\langle J_4 p,\, J_1 q \right\rangle \cdot 
\left\langle J_3 p,\, J_2 q \right\rangle
\right],
\]
yielding a forward-mode only estimator.
\end{proposition}

\textit{Proof}: \\
    Let \(A := J_1 J_2^T, B := J_3 J_4^T\), so we want to estimate \(\tr(A B)\). By the standard Hutchinson estimator \citep{hutchinson1989stochastic}, assuming \(u, v \sim H\) denote random variables as in the proposition and expand
    \begin{align*}
        \mathbb{E}\!\left[
        \left\langle J_1^\top u,\, J_2^\top v \right\rangle \cdot 
        \left\langle J_4^\top u,\, J_3^\top v \right\rangle
        \right] &= \mathbb{E}\!\left[
        \left\langle J_1^\top u,\, J_2^\top v \right\rangle \cdot 
        \left\langle J_3^\top v,\, J_4^\top u \right\rangle
        \right] \\
        &= \mathbb{E}[u^T J_1 J_2^T  v v^T J_3 J_4^Tu ] \\
        &= \mathbb{E}[u^T A  v v^T B u ]
    \end{align*}
    Marginalizing over \(u\), we have
    \begin{align*}
        \mathbb{E}[u^T A v v^T B u | u ] &= u^T A \mathbb{E}[v v^T] B u,
    \end{align*}
    using the assumptions thus yields 
    \begin{align*}
        \mathbb{E}[u^T A v v^T J_4 J_3^T u | u ] &= u^T A B u 
    \end{align*}
    So,
    \begin{align*}
        \mathbb{E}[u^T A v v^T B^T u ] &= \mathbb{E}[u^T A B^T u ] = \tr(A B).
    \end{align*}
    This proves the reverse-mode estimator. For the forward mode, we have
    \begin{align}
        \tr(J_1 J_2^T J_3 J_4^T) &= \tr(J_4^T J_1 J_2^2 J_3)
    \end{align}
    and apply the same theorem to the matrices \(\hat J_1 = J_4^T, \hat J_2 = J_1^T, \hat J_3 = J_2^T, \hat J_4 = J_3^T\), yielding
    \begin{align}
        \tr(J_1 J_2^T J_3 J_4^T) &=  \tr(\hat J_1 \hat J_2^T \hat J_3 \hat J_4^T) \\
        &=\mathbb{E}\!\left[
        \left\langle \hat J_1^\top p,\, \hat J_2^\top q \right\rangle \cdot 
        \left\langle \hat J_4^\top p,\, \hat J_3^\top q \right\rangle
        \right] \\
        &=\mathbb{E}\!\left[
        \left\langle J_4 p,\, J_1 q \right\rangle \cdot 
        \left\langle J_3 p,\, J_2 q \right\rangle
        \right]
    \end{align}
    yielding the forward-mode estimator. Note implicitly here that the domains are different: \(p, q\) are samples from the parameter space \(P\) while \(u, v\) are samples from the \(H\) space $_\square$  \\

From this, we can estimate the NTK norm,
\begin{align}
    \|\NTK\|_F^2 &= \tr(\NTK \cdot \NTK^T) \\
    &= \mathbb{E}\!\left[ \langle \vjp(u), \vjp(v) \rangle^2\right] = \mathbb{E}\!\left[ \langle \jvp(p), \jvp(q) \rangle^2\right],
\end{align}
where the \(u, v\) form is the reverse-mode estimate and the \(p, q\) form is the forward-mode estimate.
 
Likewise, the alignment, using the notation \(\NTK_i(v) = \jvp_i(\vjp_i(v))\) for \(i=1,2\), Proposition 1 yields the following expression 
\begin{align}
    \cos(\NTK_1, \NTK_2) &= \frac{\mathbb{E}\!\left[ \langle \vjp_1(u), \vjp_1(v) \rangle \langle \vjp_2(u), \vjp_2(v) \rangle\right]}{\sqrt{\mathbb{E}\!\left[ \langle \vjp_1(u), \vjp_1(v) \rangle^2 \right]\mathbb{E}\!\left[ \langle \vjp_2(u), \vjp_2(v) \rangle^2 \right]}} \\
    &= \frac{\mathbb{E}\!\left[ \langle \jvp_2(p), \jvp_1(q) \rangle^2\right]}{\sqrt{\mathbb{E}\!\left[ \langle \jvp_1(p), \jvp_1(q) \rangle^2 \right]\mathbb{E}\!\left[ \langle \jvp_2(p), \jvp_2(q) \rangle^2 \right]}}
\end{align}

Finally, the effective rank is given by 
\begin{align}
    r_{eff}(\NTK) &= \frac{\mathbb{E}[ \|\vjp(u)\|^2_2]}{\mathbb{E}\!\left[ \langle \vjp(u), \vjp(v) \rangle^2\right]} =  \frac{\mathbb{E}[ \|\jvp(p)\|^2_2]}{\mathbb{E}\!\left[ \langle \jvp(p), \jvp(q) \rangle^2\right]}
\end{align}

We found that the one-sided trace estimator could be faster than Hutch++ for trace predictions with low accuracy (around 99\%)  However, we find for now that the one-sided product estimators detailed here are typically slower than Hutch++ in all cases. Indeed, they rely on sampling independently \(u, v \sim H\) or \(p, q \sim P\) without any information about the NTK column space. Incorporating information about the NTK into these estimates, as in Hutch++, could be a future direction for improving the runtime of the one-sided estimators. We provide these here for future reference and directions.

\section{NTK for the Output of a Deep MLP}
\label{sec:apmlp}

\begin{figure}[tp]
    \centering
    \includegraphics[width=\linewidth]{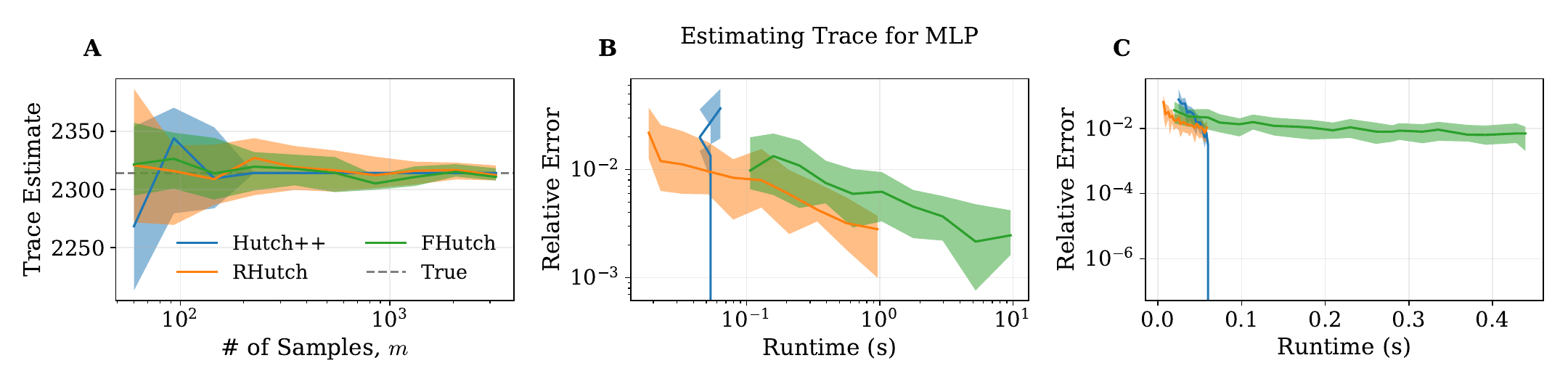}
    \caption{\textbf{Estimating \(\tr(\NTK)\) when \(h(\theta)\) tracks the output of an MLP neural network.} \textbf{A} illustrates the trace estimates for the three approaches in \sectn{methods}: Hutch++ (Algorithm~\ref{alg:hutchpp_ntk}) and the one-sided estimator (Algorithm~\ref{alg:oneside_trace}) with forward- and backward-AD (FHutch and RHutch, respectively). \textbf{B} shows the relative error \(\frac{|\tr(\NTK) - t_{m}|}{|\tr(\NTK)|}\) of each estimate, \(t_m\) versus runtime in seconds. For reference, exactly computing the trace took 1.96 seconds. \textbf{C} zooms into the low-runtime region of B and displays results with linearly scaled runtime, showing that FHutch is the slowest for this setup. Note Hutch++ obtained relative error below \(10^{-6}\) in about 0.06 seconds (about \(33\times\) faster than the exact trace). }
    \label{fig:mlp}
\end{figure}

\paragraph{Setup} While the main focus of our paper is on the generalized NTK describing updates in a model's state space, here we give an example in the classical NTK framework, in which the NTK is computed for updates in a model's outputs \citep{jacot2018neural}. Here, we use a deep Multi-Layer Perceptron (MLP) neural network. We use \(15\) layers with \(64\) hidden units per layer. As input to the network, we feed 50 distinct Gaussian 100-dimensional inputs into the first layer of the MLP, \(x \sim \mathcal{N}(0,1)^{100}\). This simple setup is chosen primarily to measure the validity of the trace estimators as runtime versus error, compared to the direct approach of computing the exact NTK trace with \(n\) matvecs. In this MLP case, the state \(h(\theta)\) reflects all activations of the output layer of the network, hence \(n = 3,200 = 50 \cdot 64\). The parameter count, \(\dim P\), is \(64,704\) in this case.   

\paragraph{Results} Fig.~\ref{fig:mlp} summarizes our results for this architecture. RHutch and FHutch denote the one-sided trace estimator in Algorithm~\ref{alg:oneside_trace} using \(G = \vjp\) and \(G = \jvp\), respectively. Evaluating the full trace with \(3,200\) matvecs required around 1.96 seconds. By contrast, the Hutch++ estimator achieved relative error below $10^{-6}$ in about 0.06 seconds, about \(33\times\) faster, using about \(600\) matvecs. Interestingly, for small sample counts, \(m < 100\), the RHutch estimator using reverse-mode AD achieved the best accuracy in the fastest time. In this typical deep-learning setup, the parameter count is much higher than the state count. Thus, in this case, RHutch samples a smaller space (\(H\)) than \(\jvp\) (\(P\)). In many applications of the NTK, estimates of the trace with relative error \(10^{-2}\) (\(99\%\) accurate) are acceptable, so using the one-sided estimator could provide a fast way to compute the NTK trace in this circumstance. However, if high accuracy is desired, we find the Hutch++ estimator always performs best.  Finally, note in this case that the forward-mode estimator was slowest because it samples in the full parameter space, \(P\), not the MLP output space \(H\). 

\paragraph{Summary} To summarize, if high accuracy is required, Hutch++ is the fastest. RHutch can be faster in the low-accuracy regime provided \(\dim(H) < \dim(P)\). By contrast, FHutch can be faster in this regime when \(\dim(P) < \dim(H)\). The next experiment illustrates the latter case.

\section{Pythia Experiments and the Transformer State-Space NTK}
	\label{app:pythia}
	\subsection{Models and Background}
	
Here, we provide further details and background on the Transformer experiments in Section~\ref{sec:transformer}. The goal is to compare the present randomized NTK estimators to direct trace computation on realistic Transformer state spaces. We use the released models
\path{EleutherAI/pythia-14m} and \path{EleutherAI/pythia-410m-deduped}, both at
step 143000, from the Pythia suite \citep{biderman2023pythiasuiteanalyzinglarge}. Using two checkpoints from the same released model family provides a controlled comparison across substantially different state and parameter dimensions.
Both models use GPT-NeoX Transformer blocks \citep{black2022gptneox20bopensourceautoregressivelanguage}.

Each Transformer block contains causal self-attention and a feed-forward network,
	with normalization and residual connections  Table~\ref{tab:pythia3_dimensions}
	lists the architecture dimensions. The models are evaluated in
	\texttt{eval} mode with key/value caching disabled. Float32 is used for both the parameters and automatic differentiation. 
	
	We utilize the statement ``Neural tangent kernels describe how parameter
	updates change the output of a neural network.'' as the fixed prompt for these Pythia Transformer experiments. We tokenize without special
	tokens and use the first 16 token IDs. We repeat the tokenized prompt if a larger sequence of tokens is requested. An all-ones attention mask is used, and GPT-NeoX
	supplies the causal mask. All tokens enter one forward pass: this experiment
	does not autoregressively generate 16 new tokens. No labels, train-test splits,
	or task loss are required to evaluate the trace at this fixed input.
	
	\begin{table}[htbp]
		\centering
		\small
		\begin{tabular}{lrr}
			\toprule
			 & Pythia-14M & Pythia-410M-deduped\\
			\midrule
			Blocks $L$ & 6 & 24\\
			Hidden width $d$ & 128 & 1,024\\
			Attention heads & 4 & 16\\
			Feed-forward width & 512 & 4,096\\
			Sequences $B$ / token positions $T$ & $1\,/\,16$ & $1\,/\,16$\\
			State dimension $n=LBTd$ & 12,288 & 393,216\\
			Differentiated parameters $p$ & 7,628,800 & 353,822,720\\
			\bottomrule
		\end{tabular}
		\caption{Dimensions of the numerical validation experiment where all blocks and 16 tokens are used.
			The parameter count excludes the vocabulary output projection.}
		\label{tab:pythia3_dimensions}
	\end{table}
	
	\subsection{The tensor-valued state, the Jacobian, and the NTK}
	\label{app:Pythia_state}
	Let $H^{(\ell)}(\theta;x)\in\mathbb R^{B\times T\times d}$ denote the hidden representation returned for Transformer block \(\ell\). In our code, by default, we stack the returned representations, excluding the initial embedding representation. Thus, we have 
	\begin{align}
		\mathcal H(\theta;x)&=\operatorname{stack}_{\ell=1}^{L}H^{(\ell)}(\theta;x)
		\in\mathbb R^{L\times B\times T\times d},
        \\
		h(\theta;x)&=\operatorname{vec}\mathcal H(\theta;x)\in\mathbb R^n.
		\label{eq:pythia3_state}
	\end{align}
    
	Before vectorization, the state is a four-dimensional tensor. We can also use a different ordering for the axes. The intermediate returned hidden states are block outputs. The final returned hidden state includes
	the final layer normalization  ``All blocks" refers to all the hidden states at the end of each transformer block and not every attention layer or feed-forward output. 
	The saved benchmark stores all 16 token positions. We do not measure vocabulary
	logits or token probabilities.
	
	The input \(x\) is held fixed across all estimator repeats for a given model, with parameter scope
	\texttt{all}, $\theta$ contains the input embedding table, all the Transformer
	blocks, and the final layer-normalization parameters.  The vocabulary projection
	is excluded because the chosen hidden state does not depend
	on it.  This is why $p$ is smaller than the nominal model size.  Including
	a parameter in this scope does not guarantee a non-zero derivative on the
	chosen sentence.  We can change the selected state blocks and the differentiated parameter scope.
After flattening the selected parameter tensors, we obtain 
	\begin{equation}
		J(\theta;x)=\frac{\partial h(\theta;x)}{\partial\theta}
		\in\mathbb R^{n\times p},\qquad \NTK(\theta;x)=JJ^\top.
		\label{eq:pythia3_kernel}
	\end{equation}
	The unflattened Jacobian has four axes, and one flattened
	parameter axis.  For two state coordinates $a=(\ell,b,t,c)$ and
	$a'=(\ell',b',t',c')$, we have
	\begin{equation}
		\NTK_{a,a'}=\sum_{j=1}^{p}
		\frac{\partial\mathcal H_a}{\partial\theta_j}
		\frac{\partial\mathcal H_{a'}}{\partial\theta_j}.
		\label{eq:pythia3_entries}
	\end{equation}

	The NTK includes cross-token and cross-layer blocks.  The trace measures the total squared parameter sensitivity of the selected hidden states, but does not compute the alignment between different state coordinates’ parameter gradients. 
	
	\subsection{Trace computation}
	 
	This subsection details the exact trace computation and randomized estimates.  The NTK is detailed in Section \ref{sec:background}. The exact trace used the following formula and did not require constructing the Jacobian or NTK.
	\begin{align*}
		\tr(\NTK)=\sum_{i=1}^{n} e_i^T\NTK e_i= ||J||_F^2
	\end{align*}
	The implementation used VJPs and JVPs to compute each summand in the summation above.  There is one complete evaluation per model.  The saved runs use chunks of 8 basis vectors and \texttt{torch.vmap} within each chunk.  While the products are computed in float32, the summation occurs in float64. \\
	The estimators are detailed in Section~\ref{sec:methods}.
	
	For Hutch++ \citep{meyer2021hutchoptimalstochastictrace},
	the more general notation is $S\in\mathbb R^{n\times q}$ and $G\in\mathbb R^{n\times r}$. Define
	\begin{equation}
		Q=\operatorname{orth}(\NTK S),\quad V=(I-QQ^\top)G,\quad
		\widehat t_{++}=\tr(Q^\top\NTK Q)+\frac1r\tr(V^\top\NTK V).
		\label{eq:pythia3_hutchpp}
	\end{equation}
	For the Pythia runs, the command-line budget \(m\) is split as \(q=\lfloor m/6\rfloor\) and \(r=\lfloor(m-q)/2\rfloor\). Thus \(m\in\{6,12,24,48,96\}\) corresponds to \(2q+r\in\{4,9,18,36,72\}\) actual NTK actions, each comprising a VJP and a JVP. 
	
	Here $q$ is the number of sketch columns, $r$ the number of residual probes,
	and $Q$ the resulting orthonormal basis matrix.  They are distinct quantities.
	The pseudocode \ref{alg:hutchpp_ntk} uses the valid equal allocation $q=r=m/3$, where we assume $m$ is divisible by 3.  It uses $2q+r=m$ NTK products and a
	residual multiplier $1/r=3/m$.  Equal allocation is not required for the
	Hutch++ decomposition.  For example, if $m=24$, we use $q=r=8$ in the pseudocode and $q=4, r=10$ in the code. 
	The floor operations allow arbitrary command-line budgets; we do not claim that this split is optimal. \\
	
	\paragraph{One-sided estimators and their sample counts.}
	The Pythia one-sided estimators use independent Rademacher probes defined as
	\begin{align}
		\widehat t_R&=\frac1m\sum_{i=1}^{m}\|J^\top z_i\|_2^2,
		&z_i&\in\mathbb R^n, \label{eq:pythia3_rhutch}\\
		\widehat t_F&=\frac1m\sum_{i=1}^{m}\|Jw_i\|_2^2,
		&w_i&\in\mathbb R^p. \label{eq:pythia3_fhutch}
	\end{align}
	The identities $\mathbb E[zz^\top]=I_n$ and
	$\mathbb E[ww^\top]=I_p$ give expectation $\tr(JJ^\top)$ in both cases.
	Here, $m$ is the actual number of independent directions. RHutch uses $m$ VJPs, and FHutch uses $m$ JVPs.  These saved runs use plain one-sided
	estimators without QR, a basis $Q$, or a split into $q$ and $r$ directions.
	Probes are processed sequentially. 
	
	For comparison to the pseudocode \ref{alg:oneside_trace}, $q$ denotes the number of orthonormal basis columns in $Q$, and $r$
	the number of independent residual probes.  The total budget
	is $m=q+r$ for the derivative calls, not $2q+r$.  Computing $Q=\operatorname{orth}(S)$ requires no
	Jacobian.  The pseudocode provided uses $q=r=m/2$ ($m$ is assumed to be even)
	and averages the residual with $1/r=2/m$.  The code
	estimators above average all $m$ squared norms with $1/m$.  Thus, the equal-split one-sided pseudocode is not the exact implementation used for these Pythia measurements. 
	
	\subsection{Implementation details and Additional Results}
	
		Each method/budget pair is run 20 times, giving 300 estimates per model.  The model, prompt, state, and parameter scope stay fixed. Only the probes change. Seeds are offsets from the base seed 20260825 and are
	stored. Model loading, tokenization, and similar things are
	excluded from timings. Hutch++ includes QR and residual projection. Initial state-space probe
	generation and exact-basis construction are not timed.
	FHutch's streamed parameter-probe generation is counted as part of the time. The randomized methods use sequential probe actions, unlike the chunk-vmapped exact reference.
	
	We plot the relative error as
	$|t_m-t_\star|/t_{\star}$ for an estimate $t_m$. Trace/error curves show medians and interquartile intervals across probe repeats. 
	For each repeat, we check the sample budgets from smallest to largest. We find where the error first reaches the chosen tolerance. We estimate the time needed by drawing a straight line between the two nearby measurements. If the smallest budget already meets the tolerance, we use $(\text{time},\text{error})=(0,1)$ as the starting point.
	We then choose the fastest method that meets the tolerance in that repeat. Repeats where no method succeeds are left out. The speedup is the exact computation time divided by this estimated time.
	This comparison is done after the measurements are collected. Here, accuracy refers to the estimated NTK trace.

    \paragraph{Numerical results}
The exact state-basis computation gives
$\tr(\NTK)=1.6038\times10^{10}$ in 101.06 seconds for Pythia-14M and
$\tr(\NTK)=2.2869\times10^{10}$ in 8,328.58 seconds for Pythia-410M-deduped.
For Pythia-14M, Hutch++ has median relative errors of 3.05\% at $m=24$ and
0.65\% at $m=96$, with median runtimes of 0.51 and 1.99 seconds, respectively;
the median fastest-method speedup at the 99\% accuracy target is $347\times$.
For Pythia-410M, at $m=24$ the median relative errors are 2.34\%, 1.66\%, and
2.08\% for Hutch++, RHutch, and FHutch, with runtimes of 2.06, 0.66, and
2.23 seconds.  At $m=96$, Hutch++ reaches 0.58\% median relative error in
8.03 seconds.  The median fastest-method speedup at 99\% accuracy is
$13{,}499\times$.

	\begin{figure}[t]
		\centering
		\includegraphics[width=\linewidth]{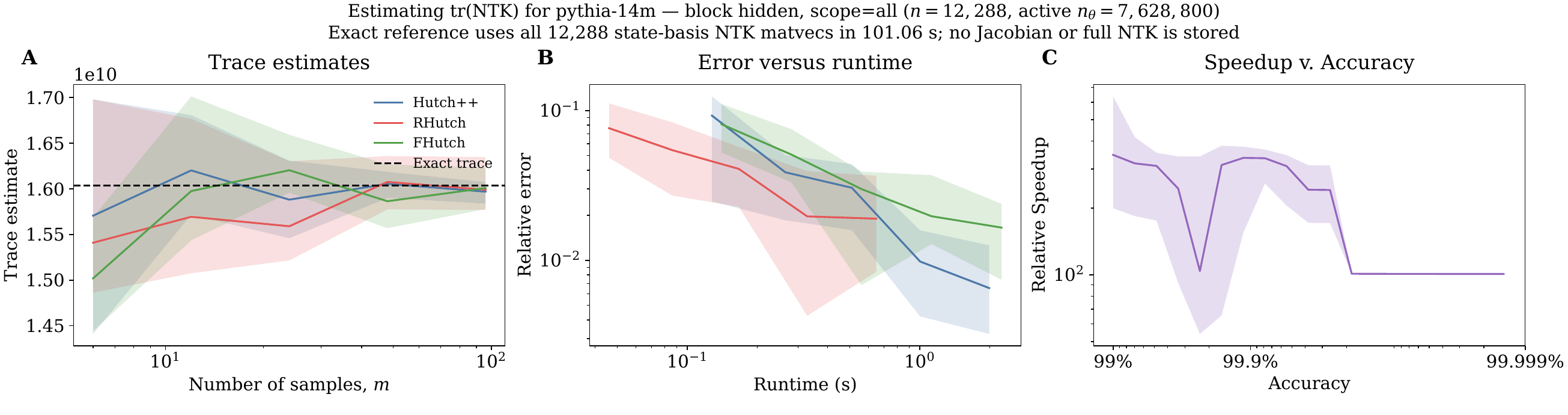}
		\caption{\textbf{Matrix-free estimation of $\tr(\NTK)$ for the hidden-state representations of released Pythia Transformers.} For each model, $h(\theta)$ contains the hidden representations of all 16 tokens at the output of every Transformer block \textbf{A} shows trace estimates against randomized sample count, $m$. \textbf{B} shows relative error $|t_m-\tr(\NTK)|/|\tr(\NTK)|$ against runtime.
			\textbf{C} shows speedup of randomized trace estimation relative to the exact state-basis evaluation. At each target accuracy and repeat, we record the fastest measured Hutch++, RHutch, or FHutch estimate satisfying the target tolerance and divide the exact-computation runtime by the estimator runtime. The most stringent-accuracy tail points should be interpreted cautiously when only a small number of estimates satisfy the target tolerance. Curves show medians over 20 estimates, and shaded regions show the $25^{\mathrm{th}}$ and $75^{\mathrm{th}}$ percentiles. } 
		\label{fig:transformer_14m_trace_exact}
	\end{figure}
	
\section{Rich and Lazy Recurrent Learning Additional Details}
\label{app:ranked_rnn}

Following \citet{liu2024how}, we trained bias-free 300-unit RNNs on row-wise sequential MNIST, where each image is presented as 28 vectors \(x_t\in\mathbb{R}^{28}\):
\begin{align}
h_0 &= 0,\qquad
h_t=e^{-1}h_{t-1}+(1-e^{-1})
\left[W_{\rm rec}\operatorname{ReLU}(h_{t-1})+W_{\rm in}x_t\right],\\
y &= W_{\rm out}\operatorname{ReLU}(h_{28}).
\end{align}
We used 55,000 training and 5,000 validation examples and trained for 10,000 updates with cross-entropy and SGD (learning rate \(0.003\), momentum \(0.9\), batch size 200, gradient clipping at 10). We evaluated 20 initialization ranks,
\[
r\in\{5,7,10,12,15,20,25,30,35,45,60,70,80,100,125,150,175,200,250,300\},
\]
with three seeds (8021--8023), giving 60 runs and 101 checkpoints per run. Matched-performance measurements use the first checkpoint reaching 95\% accuracy on a fixed 1,024-example test subset.

For each seed, rank conditions share the same underlying Gaussian initialization. We sample
\[
A_{ij}\sim\mathcal N(0,1.5^2/N),\qquad
A=U\Sigma V^\top,\qquad
W_{{\rm rec},0}
=\frac{\|A\|_F}{\|A_r\|_F}A_r,
\]
where \(A_r=U_{:,1:r}\Sigma_{1:r,1:r}V_{:,1:r}^\top\). Thus \(W_{{\rm rec},0}\) has rank \(r\) while preserving the Frobenius norm of the full Gaussian matrix; its rank is unconstrained during training.  Input and output weights are Gaussian with standard deviations \(1/\sqrt{28}\) and \(1/\sqrt{300}\), respectively.

We measure the NTK of the final pre-readout hidden state \(h_{28}\) with respect to \(W_{\rm rec}\). Using 20 fixed class-balanced validation examples,
\[
J_s=\frac{\partial\,\operatorname{vec}(H_s)}
{\partial\,\operatorname{vec}(W_{\rm rec})},
\qquad
K_s=J_sJ_s^\top,
\]
where \(H_s\in\mathbb{R}^{20\times300}\), giving a \(6000\times6000\) matrix-free NTK. We report
\[
\operatorname{align}(K_0,K_s)
=
\frac{\operatorname{tr}(K_0K_s)}
{\sqrt{\operatorname{tr}(K_0^2)\operatorname{tr}(K_s^2)}},
\qquad
r_{\rm eff}(K_s)
=
\frac{\operatorname{tr}(K_s)^2}{\operatorname{tr}(K_s^2)}.
\]
Traces use Hutch++ with 24 probes and three repeats.  Anchor examples and estimator randomness are fixed across checkpoints within each run, and error bars show mean \(\pm\) SEM over the three model seeds.

    \section{Data-Scare Knowledge Distillation Additional Details}

    We trained MLP teachers on all \(60{,}000\) Fashion-MNIST training examples and students on fixed subsets of \(N\in\{250,500,1000\}\). All models used cross-entropy classification loss.  The regularized student minimized
\[
\mathcal{L}
=(1-\lambda)\mathcal{L}_{\mathrm{CE}}
+\lambda\bigl(1-\operatorname{align}(K_s,K_t)\bigr),
\]
where \(K_s\) and \(K_t\) are the student and teacher empirical NTK operators evaluated on the same 16-example minibatch. Alignment was estimated matrix-free using 12 Hutch++ probes; no Gram matrix was materialized.

For output-space NTKs, we used one-hidden-layer MLPs with widths 512 and 128 for the teacher and student, respectively. The teacher reached \(90.05\%\) test accuracy.  Students were trained for 200 epochs over six paired seeds, using identical subsets and initializations for CE-only and regularized runs. The best fixed coefficient across subset sizes was \(\lambda=0.7\). For hidden-space NTKs, we used two-hidden-layer MLPs with the same widths and compared the stacked layer activations after a matrix-free partial trace over the hidden-unit axis. The teacher reached \(90.44\%\) test accuracy. These students were trained for 150 epochs over three paired seeds, with \(\lambda=0.5\).

\end{document}